\documentclass[journal]{IEEEtran}

\usepackage{graphicx}
\usepackage{amsmath, amssymb, amsfonts}
\usepackage{amsthm}
\usepackage{mathrsfs}
\usepackage{algorithmic}
\usepackage{algorithm}
\usepackage{array}
\usepackage{booktabs}
\usepackage{manyfoot}
\usepackage{multirow}
\usepackage{cite}
\usepackage{color}
\usepackage{xcolor}
\usepackage{url}
\usepackage{hyperref}
\usepackage{lipsum}
\usepackage{enumitem}
\usepackage{tabularx}
\usepackage{makecell}
\usepackage{adjustbox}
\usepackage{pdflscape}
\usepackage{textcomp}
\usepackage{subcaption}
\usepackage{stfloats}
\usepackage{wrapfig}
\usepackage[T1]{fontenc}
\usepackage[utf8]{inputenc}
\usepackage{indentfirst}
\usepackage{fancyhdr}
\usepackage{lineno}
\usepackage{listings}
\usepackage{ulem}

\begin{document}
\bstctlcite{IEEEexample:BSTcontrol}


\title{Prithvi-EO-2.0: A Versatile Multi-Temporal Foundation Model for Earth Observation Applications}

\author{Daniela~Szwarcman\textsuperscript{1,$\dagger$},
        Sujit~Roy\textsuperscript{2,3, $\dagger$, $\ddagger$} (Senior Member, IEEE),
        Paolo~Fraccaro\textsuperscript{1, $\dagger$, $\ddagger$},
        \TH orsteinn~El\'i~G\'islason\textsuperscript{4},
        Benedikt~Blumenstiel\textsuperscript{1},
        Rinki~Ghosal\textsuperscript{3},
        Pedro~Henrique~de~Oliveira\textsuperscript{1},
        Joao~Lucas~de~Sousa~Almeida\textsuperscript{1},
        Rocco~Sedona\textsuperscript{5},
        Yanghui~Kang\textsuperscript{6},
        Srija~Chakraborty\textsuperscript{12},
        Sizhe~Wang\textsuperscript{7},
        Carlos~Gomes\textsuperscript{1},
        Ankur~Kumar\textsuperscript{3},
        Vishal~Gaur\textsuperscript{3},
        Myscon~Truong\textsuperscript{8},
        Denys~Godwin\textsuperscript{9},
        Sam~Khallaghi\textsuperscript{9},
        Hyunho~Lee\textsuperscript{7},
        Chia-Yu~Hsu\textsuperscript{7},
        Rohit Lal\textsuperscript{3},
        Ata~Akbari~Asanjan\textsuperscript{12},
        Besart~Mujeci\textsuperscript{12},
        Disha~Shidham\textsuperscript{12},
        \textcolor{black}{Rufai Omowunmi Balogun\textsuperscript{9}},
        Venkatesh~Kolluru\textsuperscript{3},
        Trevor~Keenan\textsuperscript{11},
        Paulo~Arevalo\textsuperscript{10},
        Wenwen~Li\textsuperscript{7},
        Hamed~Alemohammad\textsuperscript{9},
        Pontus~Olofsson\textsuperscript{2},
        Timothy~Mayer\textsuperscript{3},
        Christopher~Hain\textsuperscript{2},
        Robert~Kennedy\textsuperscript{8},
        Bianca~Zadrozny\textsuperscript{1},
        David~Bell\textsuperscript{12},
        Gabriele~Cavallaro\textsuperscript{4,5} (Senior Member, IEEE),
        Campbell~Watson\textsuperscript{1},
        Manil~Maskey \textsuperscript{2} (Senior Member, IEEE),
        Rahul~Ramachandran\textsuperscript{2},
        and Juan Bernabe Moreno\textsuperscript{1}\\
        $\dagger$Equal Contribution;

\thanks{\textsuperscript{1}IBM Research (UK, Ireland, Brazil, Zurich, and USA); \textsuperscript{2}NASA Marshall Space Flight Center, Huntsville, AL, USA; \textsuperscript{3}Earth System Science Center, University of Alabama in Huntsville, AL, USA; \textsuperscript{4}School of Engineering and Natural Sciences, University of Iceland, Reykjavik, Iceland; \textsuperscript{5}J\"ulich Supercomputing Centre, Forschungszentrum J\"ulich, J\"ulich, Germany; \textsuperscript{6}Department of Biological Systems Engineering, Virginia Tech, Blacksburg, VA, USA; \textsuperscript{7}School of Geographical Sciences and Urban Planning, Arizona State University, Tempe, AZ, USA; \textsuperscript{8}College of Earth, Ocean, and Atmospheric Sciences, Oregon State University, Corvallis, OR, USA; \textsuperscript{9}Center for Geospatial Analytics, Clark University, Worcester, MA, USA; \textsuperscript{10}Department of Earth and Environment, Boston University, Boston, MA, USA; \textsuperscript{11}Department of Environmental Science, Policy, and Management, University of California, Berkeley, CA, USA; \textsuperscript{12}Earth from Space Institute, Universities Space Research Association, USA}%

\thanks{$\ddagger$ Corresponding author: sujit.roy@nasa.gov, paolo.fraccaro@ibm.com\\
Model available at: \url{https://huggingface.co/ibm-nasa-geospatial/Prithvi-EO-2.0}. Code available at: \url{https://github.com/NASA-IMPACT/Prithvi-EO-2.0}}
}

\maketitle

\begin{abstract}
\small
This paper presents Prithvi-EO-2.0, a new geospatial foundation model that offers significant improvements over its predecessor, Prithvi-EO-1.0. Trained on 4.2 million global time series samples from NASA’s Harmonized Landsat and Sentinel-2 data archive at 30-m resolution, the new model incorporates temporal and location embeddings for enhanced performance across various geospatial tasks. Through extensive benchmarking with GEO-Bench, the model outperforms the previous Prithvi-EO model by 8\% across a range of tasks. It also outperforms six other geospatial foundation models when benchmarked on remote sensing tasks from different domains and resolutions (i.e. from 0.1 m to 15 m). The results demonstrate the versatility of the model in both classical Earth observation and high-resolution applications. Early involvement of end-users and subject matter experts (SMEs) allowed constant feedback on model and dataset design, enabling customization across diverse SME-led applications in disaster response, land cover and crop mapping, and ecosystem dynamics monitoring. Prithvi-EO-2.0 is available as an open-source model on Hugging Face and IBM \textcolor{black}{TerraTorch}, with additional resources on GitHub. The project exemplifies the Trusted Open Science approach embraced by all involved organizations.

\end{abstract}

\newpage
\section{Introduction}\label{sec:intro}
\IEEEPARstart{I}{n} recent years, Earth Observation (EO) has entered a new era with the rise of Geospatial Foundation Models (GFMs)—large-scale artificial intelligence (AI) systems trained on vast amounts of unlabeled satellite imagery using self-supervised learning~\cite{reviewEO}. Foundation models are general-purpose neural networks (often based on transformer architectures) that learn broad representations from massive datasets and can be fine-tuned for many downstream tasks with minimal labeled data\textcolor{black}{~\cite{opportunitiesrisksfoundationmodels}}. These models \textcolor{black}{can} help address long-standing challenges in EO by reducing the need for manually labeled \textcolor{black}{samples}, 
\textcolor{black}{which are usually hard to obtain at scale~\cite{reviewEO}}. Remote sensing is \textcolor{black}{especially} well-suited for foundation model development due to its global coverage, frequent revisits, and the sheer scale of unstructured imagery. Once pretrained, \textcolor{black}{GFMs} require less data to achieve similar or even \textcolor{black}{superior performance} across various domains\textcolor{black}{~\cite{satmae, presto}. Despite their potential,} real-world adoption remains limited.

We \textcolor{black}{identify three} main limitations with available GFMs. First, \textcolor{black}{although} EO data is \textcolor{black}{inherently} multi-temporal, most GFMs do not account for \textcolor{black}{this characteristic. Those} that do either \textcolor{black}{process} only point data or focus on long time series \textcolor{black}{restricted to} small patches or \textcolor{black}{individual} pixels~\cite{presto, ubarn}. 
\textcolor{black}{Second, in-depth validation considering diverse types of tasks and clear comparison protocols remains limited. This hinders users' ability to assess whether the models are suitable for their use case. Third, adapting state-of-the-art GFMs to different applications may require AI expertise in the absence of the proper tools and guidance. While several GFMs have released the weights and model architecture \cite{presto, DOFA, decur, satlas, satmae}, which is an important step toward community adoption, we believe that the lack of clear instructions or a streamlined code base for fine-tuning remains a significant barrier to broader use and further evaluation of GFMs.}

To \textcolor{black}{address} these issues and \textcolor{black}{increase} the impact of GFMs \textcolor{black}{within the EO} community, we developed a new multi-temporal GFM called Prithvi-EO-2.0\footnote{Prithvi means Earth in Sanskrit.}. \textcolor{black}{Building on} its US-only predecessor, Prithvi-EO-1.0~\cite{prithvi_v1}, \textcolor{black}{our new model} explicitly uses transformer attention in both \textcolor{black}{spatial and temporal dimensions and also incorporates metadata for location and time to better organize the embedding space}. Pretraining was \textcolor{black}{conducted} at scale on a large dataset of medium resolution (30~m) satellite imagery \textcolor{black}{created} from NASA's Harmonized Landsat Sentinel-2 (HLS) archive spanning a decade. We designed a new sampling strategy for \textcolor{black}{our} pretraining dataset to \textcolor{black}{focus on} long-term trends and seasonal patterns, while also ensuring \textcolor{black}{diverse and high-quality samples}, so the model can better capture these characteristics.
\textcolor{black}{The scale of this dataset} also allowed us to increase the model size up to 600 \textcolor{black}{million} parameters, \textcolor{black}{placing Prithvi-EO-2.0} among the largest in the field of EO.

\textcolor{black}{Additionally, we conducted} extensive validation, \textcolor{black}{in close collaboration with remote sensing subject matter experts} (SMEs)\textcolor{black}{, who were also involved in the dataset and model design}. In particular, we validated \textcolor{black}{Prithvi-EO-2.0} through a comprehensive benchmarking process with GEO-Bench~\cite{GEOBench}, a robust framework for assessing \textcolor{black}{EO} Foundation Models. Further, we evaluated the models \textcolor{black}{across a diverse set of} applications, \textcolor{black}{with} SMEs \textcolor{black}{playing a central role in implementation and assessment of results}. To facilitate adoption and fine-tuning \textcolor{black}{for new downstream tasks, we integrated} Prithvi-EO-2.0 into TerraTorch\footnote{\url{https://github.com/IBM/terratorch}}, a toolkit \textcolor{black}{powered by PyTorch Lightning\footnote{\url{https://lightning.ai/docs/pytorch/stable/}} and TorchGeo\footnote{\url{https://torchgeo.readthedocs.io/en/stable/}}} that simplifies customization of GFMs for various EO applications. \textcolor{black}{The data loaders used in this work were also included in TerraTorch for reproducibility. Our results show} that Prithvi-EO-2.0 can generalize across \textcolor{black}{a wide range of} remote sensing tasks\textcolor{black}{, spanning} different spatiotemporal resolutions and domains, \textcolor{black}{often} requiring fewer labeled samples to achieve \textcolor{black}{strong} performance \textcolor{black}{compared to baseline models}. 

The structure of the paper is as follows. First, we \textcolor{black}{introduce related work and then our methodology, starting with the description of} our sampling strategy and the processes to build our global pretraining dataset. \textcolor{black}{Next}, we \textcolor{black}{present the Prithvi-EO-2.0 architecture and the} pretraining procedure. \textcolor{black}{We then detail our evaluation approach, including benchmarking on the GEO-Bench~\cite{GEOBench} framework and SME-led downstream tasks. Finally, we present our results for benchmarking and applications, and discuss the outcomes}.


\section{\textcolor{black}{Related Work}}

\textcolor{black}{Recent advances in self-supervised learning and large-scale pretraining have enabled the development of GFMs for EO data. Early work includes SatMAE~\cite{satmae}, which adapts Masked Autoencoders (MAE) to satellite imagery with two variants: a spectral model with multiple patch embeddings grouping similar bands, and a temporal model trained on sequences of three RGB images with a shared 2D patch embeddings layer \cite{satmae}. Although multispectral and temporal aspects are studied, they are not combined into a unified model. Scale-MAE~\cite{scalemae} also builds on MAE, addressing scale variation by introducing a band-pass filter to separate and reconstruct low- and high-frequency images at different scales \cite{scalemae}. However, Scale-MAE is limited to RGB data and lacks temporal modeling.}

\textcolor{black}{Other approaches explore contrastive learning and distillation methods. The SSL4EO-S12~\cite{ssl4eos12} is a large-scale dataset of Sentinel-1/2 imagery (3M patches, four seasonal frames per location). The authors pretrain ResNet-50 and ViT-S backbones using MoCo (contrastive) and DINO (distillation) on the Sentinel-2 part \cite{ssl4eos12}. The frames, however, serve only as augmentation rather than explicit sequence modeling.}

\textcolor{black}{DeCUR~\cite{decur} and DOFA~\cite{DOFA} invest in multimodal pretraining, with the first adopting the SSL4EO-S12 dataset and the latter considering several data sources. DeCUR employs separate encoders for each modality and proposes to decouple common and unique modality representations via multimodal redundancy reduction \cite{decur}. DOFA adopts a hypernetwork approach that generates weights conditioned on the central wavelength of each band. The pretraining procedure combines masked image modeling with distillation from ImageNet-based models \cite{DOFA}. Both DeCUR and DOFA operate on 2D inputs only.}

\textcolor{black}{Temporal modeling is the focus of Presto~\cite{presto}, a lightweight transformer for EO pixel time series. The model is trained with 12 months of data from different sources, with each month represented by a monthly timestamp, including Sentinel-1/2, topography, ERA-5, and metadata. While effective for time-series tasks, Presto is evaluated on image tasks in a limited manner: nine pixels are sampled from the images, and the results are the mode of the corresponding predictions \cite{presto}. In contrast, U-BARN~\cite{ubarn} also considers the spatial aspect: a BERT-like approach is used for pretraining, with an architecture that combines a U-Net with a transformer. Despite its spatiotemporal design, U-BARN is trained with small 64~$\times$~64 patches and a very restricted pretraining dataset, considering only nine Sentinel-2 tiles from France over two years \cite{ubarn}.}

\textcolor{black}{In this context, Prithvi-EO-2.0 focuses on a larger, higher-quality global dataset with more than 4M samples, as well as larger models, and flexibility regarding the use of metadata. Our approach emphasizes spatiotemporal modeling, incorporating seasonal and long-term dynamics over a decade, significantly exceeding the temporal coverage of prior models, while maintaining spatial representation capabilities.}

\section{Methods}
\subsection{Dataset Description and Sampling}\label{sec:CONTUSsampling}

The HLS \textcolor{black}{product}~\cite{HLSpub} is a deliverable from the Satellite Needs Working Group that harmonizes data from NASA/USGS's Landsat 8 and 9 and the ESA's Sentinel-2A and -2B satellites to achieve a higher temporal resolution. HLS is compatible with the 40-year Landsat data record at 30~m spatial resolution but with a temporal resolution of two-three days on average. \textcolor{black}{The data is framed into tiles measuring 109.8 km by 109.8 km (3,660 $\times$ 3,660 pixels) in the UTM-based Military Grid Reference System (MGRS)\cite{HLSpub}}. Fifteen visible and infrared bands are available in HLS, however, not all spectral bands are present in Sentinel and Landsat. Therefore, we used the six spectral bands common to both: Blue, Green, Red, NIR, SWIR1, and SWIR2. This band combination will be referred to as the six-band HLS reflectance data.

\begin{figure}[hbt!]
\includegraphics[width=\columnwidth]{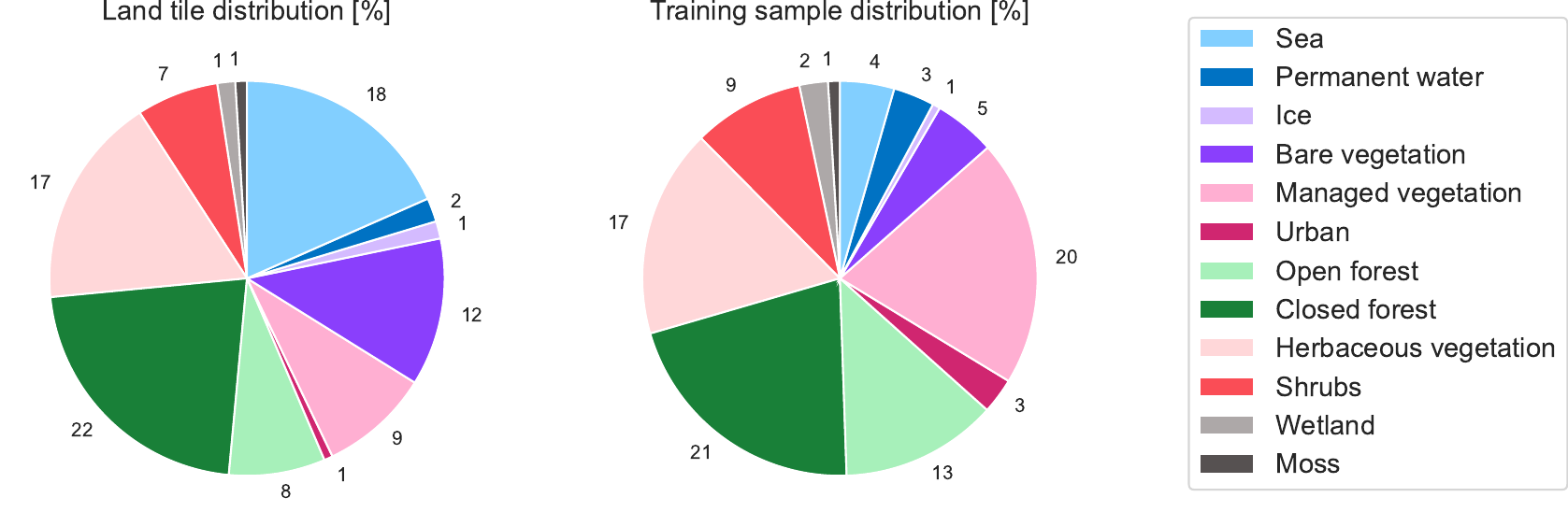}
\centering
\caption{LULC distribution of the training samples in comparison to all land tiles.}
\label{fig:lulc}
\end{figure}

Our sampling approach aimed at creating a high-quality dataset representing diverse land cover and ecosystems for robust model training, while minimizing cloud and missing data issues. To achieve this, we first calculated the proportion of LULC classes and ecoregions for each HLS tile using the Copernicus Land Cover 100~m~\cite{landcover2019} and RESOLVE Ecoregions~\cite{ecoregions2017} labels. Second, after merging the 12 closed and open forest classes into two classes, we sampled 100 tiles \textcolor{black}{(full size of 3,660 $\times$ 3,660 pixels)} per LULC class from the 500 tiles with the highest class proportion. Urban areas were over-sampled by selecting 1,000 tiles that cover about 60\% of the global urban areas. Additionally, we included 1,000 tiles with a high LULC class entropy to capture heterogeneous landscapes. Next, we ensured that the 846 ecoregions were represented if the region's size allowed it, which is threshold\textcolor{black}{ed} at 5\% of tile area coverage. In total, 712 ecoregions are present in three or more tiles, while 68 are not included due to limited area coverage. We applied a 95\%-5\% \textcolor{black}{training}-validation split and dropped 133 sampled tiles due to quality issues or insufficient data coverage, e.g., in Greenland and Antarctica.
This process resulted in 3,028 \textcolor{black}{training} and 163 validation tiles. The LULC distribution of the final dataset is visualized in Figure~\ref{fig:lulc}, showing that sea and bare vegetation (desert) classes were downsampled compared to all 18,000 HLS land tiles. Because we oversampled build-up areas, managed vegetation is also over-represented due to the geographical proximity.

Once \textcolor{black}{we} ensured that \textcolor{black}{our selection included a} diversified and representative set of tiles, the next phase of our dataset preparation aimed at optimizing both temporal and spatial coverage \textcolor{black}{when} sampling individual patches from the selected HLS tile \textcolor{black}{IDs} (i.e.\textcolor{black}{,} the actual satellite images used in pretraining). This was particularly important \textcolor{black}{to allow} Prithvi-EO-2.0 to \textcolor{black}{capture seasonal patterns and longer-term changes}. To achieve this, we \textcolor{black}{built} temporal sequences of four HLS images with \textcolor{black}{intervals of one to six months between consecutive timestamps, spanning the period from} 2014 and 2023. \textcolor{black}{By enforcing a minimum separation between consecutive timestamps, we ensure that each sample contains images spanning a longer period of time, even with a relatively small sequence. However, the most relevant pattern changes might occur at different time scales across regions, and no timestamp selection strategy is optimal for the entire globe. Our approach guarantees at least monthly variations between images, which can highlight seasonal changes while avoiding overly similar consecutive images. Sequence length is another trade-off: longer sequences increase the number of tokens to process and make pretraining more computationally expensive. Therefore, we used samples with four timestamps, as it leads to a reasonable token count while still gathering images from four different months.}

The sequences were iteratively sampled from the selected HLS tiles until all candidates (minimum \textcolor{black}{of four} timestamps) were processed or 1,500 samples per tile (250 for validation) were reached. 
Then, the sequences were split into non-overlapping 256~$\times$~256 patches with four consecutive timestamps.
We discarded samples that contained more than 1\% of missing value pixels in any band or more than 20\% of cloudy pixels using the cloud and cloud shadow data contained in the \textit{Fmask} band. \textcolor{black}{M}issing value pixels were filled \textcolor{black}{using} nearest\textcolor{black}{-neighbor} interpolation. We ensured that each \textcolor{black}{256~$\times$~256} patch did not contain more than ten samples \textcolor{black}{(two for validation)} and randomly downsampled otherwise to \textcolor{black}{avoid} over-representing cloud-free areas and \textcolor{black}{maintain} spatial diversity. 

Finally, all training samples that spatially overlapped any validation area were discarded. This patch-sampling strategy \textcolor{black}{helped} maintain high data quality while ensuring a comprehensive representation of different temporal and spatial contexts.

\begin{figure*}[htp]
\centering
\includegraphics[width=1.8\columnwidth]{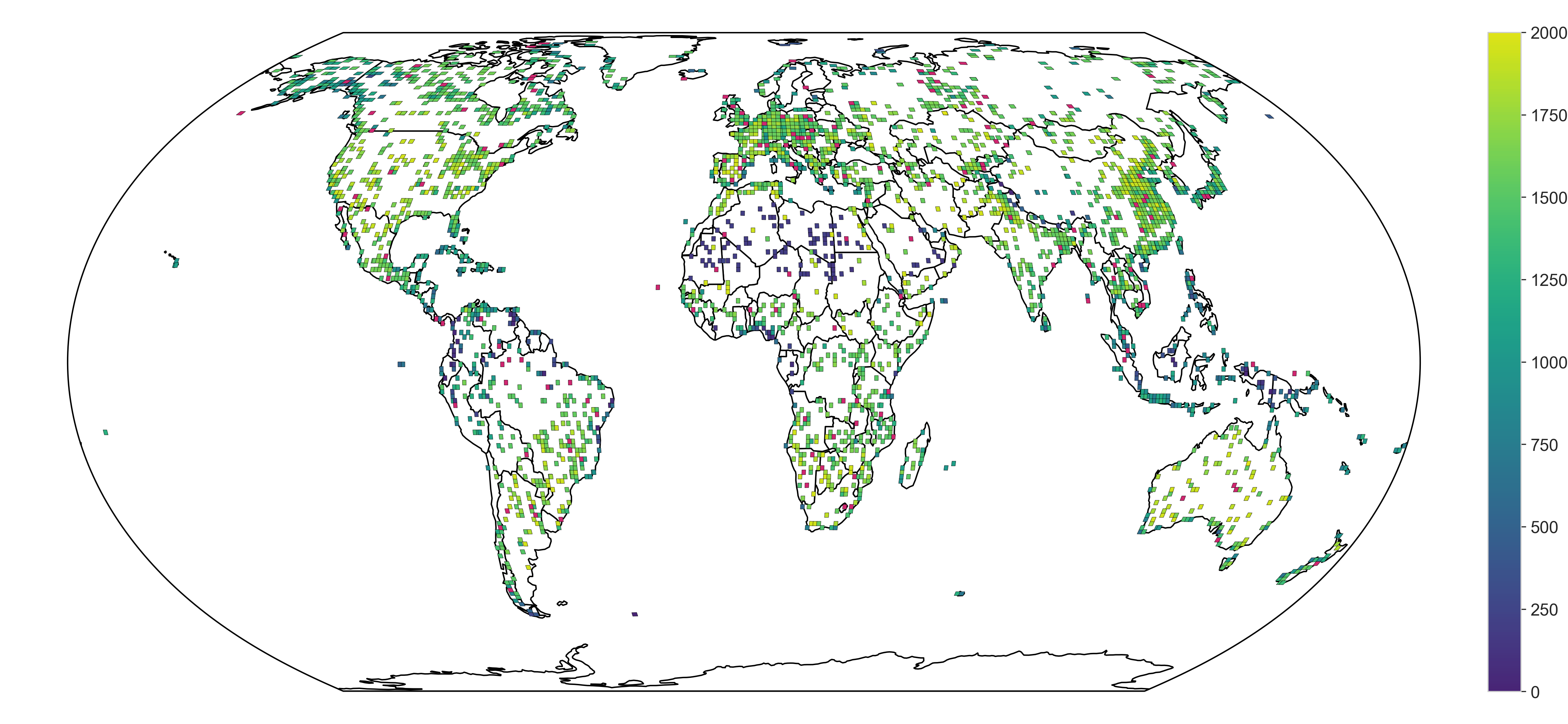}
\caption{Global HLS dataset distribution visualized on a tile-level. The number of training samples are color-coded in blue to green, while validation tiles are visualized in magenta.}
\label{fig:dataset_map}
\end{figure*}

After preliminary pretraining experiments, we applied additional filtering to address data quality issues and \textcolor{black}{keep} stable training conditions. \textcolor{black}{Specifically}, HLS tiles from central Greenland with persistent \textcolor{black}{artifacts} were removed. We also downsampled full sea and desert regions due to their very low or high reflectance values to improve pretraining stability and avoid over-representation of homogeneous areas. Sea\textcolor{black}{-only} samples were filtered using the \textit{Fmask} band, while desert samples were randomly subsampled from all desert tiles. 
The resulting pretraining dataset includes 4.2M training and 46k validation samples of size \textcolor{black}{256~$\times$~256} pixels, which are visualized in Figure~\ref{fig:dataset_map}.

\subsection{\textcolor{black}{Model architecture and pretraining}}

The pretraining process of our foundation model is based on the masked autoencoder (MAE) approach~\cite{mae}, a self-supervised learning method widely used and extended for different data types, including video~\cite{videomae} and multispectral images~\cite{satmae, scalemae}. The MAE reconstructs masked images using an asymmetric encoder-decoder architecture with a Vision Transformer (ViT) backbone~\cite{dosovitskiy2020image}. In detail, each input image is divided into non-overlapping patches of the same size and the ViT encoder embeds the patches using a linear projection with added 2D sin/cos positional embeddings. A subset of the patches is randomly masked and only the unmasked tokens are processed by the \textcolor{black}{t}ransformer blocks of the encoder~\cite{mae}. The decoder receives the encoded visible patches and mask tokens, which are learned vectors representing a missing token to be predicted. Positional embeddings are added to the full set of tokens and sent to the decoder's series of transformer blocks to perform the image reconstruction task~\cite{mae}. The loss function is the mean squared error (MSE) between the masked and predicted tokens in the pixel space~\cite{mae}.

\begin{figure*}[htp]
\centering
\includegraphics[width=\textwidth]{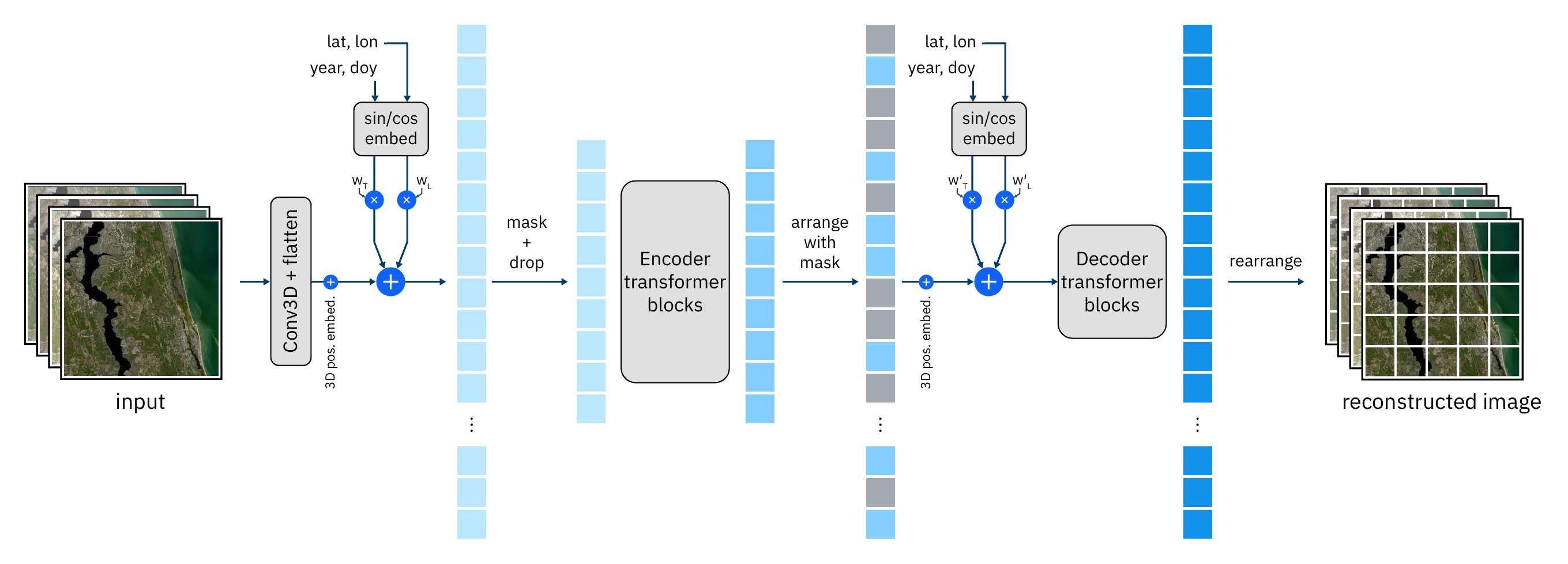}
\caption{\textcolor{black}{Prithvi-EO-2.0 architecture and general pretraining framework.}}
\label{fig:MAE}
\end{figure*}

Figure \ref{fig:MAE} shows the overall MAE structure for pretraining Prithvi\textcolor{black}{-EO-2.0}, which modifies the original MAE in two ways. First, we replaced the 2D patch embeddings and 2D positional embeddings with 3D versions to support inputs with spatiotemporal characteristics, i.e., a sequence of $T$ images of size $(H, W)$. Following approaches to adapt MAE to video processing~\cite{videomae, spatiotemporalmae}, our 3D patch embeddings consist of a 3D convolutional layer, dividing the 3D input into non-overlapping cubes of size $(t, h, w)$ for time, height, and width dimensions, respectively. \textcolor{black}{These are image dimensions: number of images in the sequence and height and width in pixels}. Unlike videos, satellite data can be acquired at non-regular and relatively low frequency in time (days), so we set $t = 1$ in our models. For the 3D positional encodings, we first generate 1D sin/cos encodings \textcolor{black}{\cite{mae, att_is_all_you_need}} individually for time, height, and width dimensions and then combine them into a single 3D positional encoding. \textcolor{black}{The 1D sin/cos positional encodings are defined as \cite{att_is_all_you_need}:}

\begin{equation}
\begin{aligned}
\textcolor{black}{\mathrm{PE}_{pos,2i} = \sin\left( \frac{pos}{10000^{2i / D}} \right)} \\[2pt]
\textcolor{black}{\mathrm{PE}_{pos,2i+1} = \cos\left( \frac{pos}{10000^{2i / D}} \right)}
\end{aligned}
\label{eq:posenc}
\end{equation}

\noindent where $pos$ represents the position, $i$ is the dimension, and $D$ is the embedding dimension size.

Second, we considered metadata on geolocation (latitude and longitude) and date (year and day-of-year ranging 1-365) in pretraining. The model receives time and location information for each sample \textcolor{black}{(i.e., date for each image in the sequence and the center latitude and longitude) and encodes them independently. Latitude and longitude are encoded separately using the 1D sin/cos strategy, similar to equation~\ref{eq:posenc}, and the resulting embeddings are concatenated together. The same is done for year and day-of-year.}

\textcolor{black}{The metadata encodings} are added to the embedded tokens via a weighted sum with learned weights: one for time and one for location (see Figure~\ref{fig:MAE}). This means that we let the model learn how the metadata should be added to the embedded tokens. Since this metadata is often not available, we decided to include it not as additional inputs in patch embeddings, but as a simple bias, similar to the idea of positional encodings. For the same reason, we also added a drop mechanism during pretraining that randomly drops the geolocation and/or the temporal data to help the model learn how to handle the absence of this information. \textcolor{black}{Note that the metadata encodings are designed to inform the model about the sample's general location (center coordinates) and corresponding timestamp, while only the 3D positional encodings are responsible for conveying the positions of the tokens within the sample.}

\textcolor{black}{Using the architecture and MAE approach described}, we developed Prithvi-EO-2.0 in two sizes, 300M and 600M, based on ViT-L and ViT-H, respectively~\cite{dosovitskiy2020image}. We trained versions with (Prithvi-EO-2.0-$\ast$-TL) and without (Prithvi-EO-2.0-$\ast$) temporal (T) and location (L) information. The models were trained for 400 epochs \textcolor{black}{on A100 40GB GPUs}; \textcolor{black}{for the 300M models we used 80 GPUs} (utilizing $\sim$21,000 GPU-hours for the full training), while \textcolor{black}{for the 600M models,} 240 GPUs (utilizing $\sim$58,000 GPU-hours for the full training). The local batch size (i.e., the batch size per GPU) \textcolor{black}{was set to 48 for ViT-L and 16 for ViT-H}, to accommodate the larger model; a global batch size (local batch size multiplied by the number of GPUs) of 3,840 was used for all models. \textcolor{black}{We used a weight decay of} 0.05 and a cosine scheduler with a maximum learning rate of 5e-4 and linear warmup for 40 epochs starting from a learning rate of 1e-6. For models with temporal and location data, we set the drop probability to 0.1. As described in section \ref{sec:CONTUSsampling}, the samples have shape \textcolor{black}{4~$\times$~256~$\times$~256} for time, height, and width, respectively, and we used random crop \textcolor{black}{to 4~$\times$~224~$\times$~224} and random horizontal flips as data augmentation. The models were pretrained on the Booster partition of the JUWELS (Jülich Wizard for European Leadership Science) supercomputer~\cite{JUWELS} hosted at JSC.

\subsection{\textcolor{black}{Evaluation}}

There are two key aspects \textcolor{black}{to consider} when evaluating the usefulness of a foundation model. The first \textcolor{black}{involves benchmarking the model against} published competitors \textcolor{black}{using a standardized} set of tasks, following a rigorous protocol that \textcolor{black}{ensures fair comparison} and reproducibility. The second \textcolor{black}{focuses on assessing the model’s performance} in real-world downstream tasks, \textcolor{black}{comparing it to state-of-the-art methods in those domains}.

\textcolor{black}{For the benchmarking evaluation, we selected GEO-Bench \cite{GEOBench}, a widely used} and rigorous benchmark framework available for \textcolor{black}{EO} foundation models. 

\textcolor{black}{As for the downstream evaluation, SMEs assessed the Prithvi-EO-2.0 models across multiple tasks in three categories of applications: \textit{disaster response}, \textit{land cover and crop mapping}, and \textit{ecosystem dynamics}.} The approaches adopted in each use case varies because of the different \textcolor{black}{requirements} and practices in \textcolor{black}{their respective} domains, reinforcing the importance of testing the model with a variety of real-world applications. \textcolor{black}{The overall characteristics of the downstream tasks can be seen in} Table~\ref{tab:downstreamtasks}.



\textcolor{black}{For} classification tasks, we \textcolor{black}{use accuracy as the evaluation metric, defined as}:
\begin{equation}
\text{Accuracy}=\frac{TP + TN}{TP+TN+FP+FN},
    \label{eq:acc}
\end{equation}
\textcolor{black}{where TP, TN, FP and FN denote the number of true positives, true negatives, false positives and false negatives respectively.} 
For segmentation tasks, we \textcolor{black}{use the mean Intersection over Union (mIoU) as the evaluation metric. It measures the average overlap between predicted and ground truth regions across all classes, defined as:}
\begin{equation}
\text{mIoU}=\frac{1}{C}\sum_{c}^{C}\frac{prediction_{c}\,\cap truth_{c}}{prediction_{c}\,\cup truth_{c}},
    \label{eq:miou}
\end{equation}
\textcolor{black}{where $C$ is the total number of classes and $prediction_{c}$ and $truth_{c}$ denote the predicted and ground truth regions for class $c$, respectively. For some tasks, we also report} the precision, recall, and F1-score, defined as:

\begin{align}
\text{Precision} &= \frac{TP}{TP + FP} \\
\text{Recall} &= \frac{TP}{TP + FN} \\
\text{F1\text{-}Score} &= \frac{2TP}{2TP + FP + FN}
\end{align}

\textcolor{black}{In the following subsections, we provide a detailed description of the benchmarking protocol and the downstream tasks for each of the three application categories (disaster response, land cover and crop mapping, and ecosystem dynamics). We highlight that all fine-tuning experiments conducted in this work require a single GPU.}

\subsubsection{Benchmarking}

\textcolor{black}{GEO-Bench offers a carefully curated collection of six classification and six semantic segmentation datasets \cite{GEOBench}, encompassing a range of spatial resolutions and domains (see Table~\ref{tab:dataset_characteristics}). Additionally, it establishes an evaluation protocol \cite{GEOBench} designed to enable fair and reproducible comparisons across models. This protocol begins with hyperparameter tuning, where each model is allowed a fixed number of trials (ten in our study) on each dataset. The best hyperparameters identified on the validation set for each task are then used to conduct repeated experiments (N = 10) with different random seeds. Repeating the experiments accounts for the inherent randomness in training AI models, thereby enabling more robust and reliable comparisons of results.}

\begin{table}[htbp]  
\centering
\caption{\textcolor{black}{Characteristics of the GEO-Bench datasets~\cite{GEOBench}. We show image height and width sizes in pixels (H,W), number of classes (C), sizes of the training (Train), validation (Val), and test (Test) sets, numbers of bands (B), and sensors (Sensors). S2 and S1 refers to Sentinel-2 and Sentinel-1, respectively.}}
\label{tab:dataset_characteristics}
\small
\setlength{\tabcolsep}{3.2pt} 
\renewcommand{\arraystretch}{1.4} 
\begin{tabular}{@{}lccccccc@{}}
\toprule
\multicolumn{8}{c}{\textbf{Classification}} \\ \midrule
\textbf{Name} & \textbf{H,W} & \textbf{C} & \textbf{Train} & \textbf{Val} & \textbf{Test} & \textbf{B} & \textbf{Sensors} \\ \midrule
m-bigeartnet & 120 & 43 & 20,000 & 1,000 & 1,000 & 12 & S2 \\
m-so2sat & 32 & 17 & 19,992 & 986 & 986 & 18 & S2 + S1 \\
m-brick-kiln & 64 & 2 & 15,063 & 999 & 999 & 13 & S2 \\
m-forestnet & 332 & 12 & 6,464 & 989 & 993 & 6 & Landsat-8 \\
m-eurosat & 64 & 10 & 2,000 & 1,000 & 1,000 & 13 & S2 \\
m-pv4ger & 320 & 2 & 11,814 & 999 & 999 & 3 & RGB \\ \midrule
\multicolumn{8}{c}{\textbf{Segmentation}} \\ \midrule
\textbf{Name} & \textbf{H,W} & \textbf{C} & \textbf{Train} & \textbf{Val} & \textbf{Test} & \textbf{B} & \textbf{Sensors} \\ \midrule
m-pv4ger-seg & 320 & 2 & 3,000 & 403 & 403 & 3 & RGB \\
m-chesapeake & 256 & 7 & 3,000 & 1,000 & 1,000 & 4 & RGBN \\
m-cashew-plant & 256 & 7 & 1,350 & 400 & 50 & 13 & S2 \\
m-SA-crop-type & 256 & 10 & 3,000 & 1,000 & 1,000 & 13 & S2 \\ 
m-nz-cattle & 500 & 2 & 524 & 66 & 65 & 3 & RGB \\ \addlinespace[3pt]
m-NeonTree & 400 & 2 & 270 & 94 & 93 & 5 & \makecell{RGB \\+ Hypersp. \\+ Elevation} \\ \bottomrule
\end{tabular}
\end{table}

\textcolor{black}{We added the GEO-Bench datasets} to TerraTorch \textcolor{black}{and used} its companion library TerraTorch-iterate\textcolor{black}{\footnote{\url{https://github.com/IBM/terratorch-iterate}}} to perform the hyperparameter tuning search via Bayesian Optimization in Optuna\footnote{\url{https://optuna.readthedocs.io/en/stable/}}, as well as the repeated experiments. We compare Prithvi-EO-2.0 against six of the most popular and performant EO foundation models recently released \textcolor{black}{for optical data} (see Table~\ref{tab:GFMs_models}), as well as the earlier version of Prithvi trained \textcolor{black}{with data only from the US} released last year~\cite{prithvi_v1}. For classification tasks, all models were paired with a linear layer, while we used a U-Net architecture~\cite{unet} for ResNets and UPerNet~\cite{upernet} decoder for transformer-based architectures, which represent the most adopted choices for these types of architectures. Hyperparameter tuning was done across the same search space that included learning rate, decoder depth, and weight decay. For comparability, batch size was fixed to a reasonable value for each dataset for all models. Finally, we used only the multispectral sensors from each dataset to conduct our benchmarking experiments. For each model, we utilized the specific spectral bands it was pretrained on. \textcolor{black}{For example, if a model was pretrained on RGB images and the dataset contains all Sentinel-2 bands, only the RGB bands were provided. We use the same data augmentation techniques (random horizontal and vertical flips) for all models and datasets. Also, all images} were uniformly resized to 224~$\times$~224 pixels.

To conduct the benchmark comparison, we computed \textcolor{black}{the mIoU for the segmentation tasks and the overall accuracy (micro-averaged accuracy) for the classification tasks; the exception is the \textit{m-big-earthnet} dataset, which is a multi-label classification task, and we used the micro-averaged F1 score}. For each GFM evaluated, we averaged those metrics to obtain an aggregated measure of each model's performance, grouping them in classification, segmentation, and combined tasks. While \textcolor{black}{these} metrics are different in nature, they range between 0 and 1. Therefore, we believe this procedure is appropriate to get an overall performance picture across GEO-Bench datasets. 

\begin{table*}[htbp]
\centering
\caption{Characteristics of the various GFMs compared to Prithvi-EO-2.0: \textcolor{black}{backbone type (Type)}, number of parameters (\# Param.), pretraining technique (Technique) and the data used in pretraining (Data), its resolution (Res.), number of samples (N), and number of \textcolor{black}{timestamps} in a sample (T). Characteristics reported for the Satlas model pretrained on Sentinel-2 downstream tasks. Sample size of pretraining dataset (i.e. N column) is estimated when not clearly reported.}
\label{tab:GFMs_models}
\small
\begin{tabularx}{1.8\columnwidth}{@{}Xc@{\hspace{20pt}}c@{\hspace{20pt}}c@{\hspace{30pt}}c@{\hspace{30pt}}c@{\hspace{20pt}}c@{\hspace{20pt}}c@{}}
\toprule
\textbf{Model} & \textbf{Type} & \textbf{\# Param.} & \textbf{Technique} & \textbf{\textcolor{black}{Data}} & \textbf{Res.} & \textbf{N} & \textbf{T} \\
\midrule
MOCO~\cite{ssl4eos12} & ResNet50 & 25M & \makecell[t]{Contrastive\\learning} & \mbox{Sentinel-2} & 10m & 1M & 1 \\ \addlinespace[2mm]
DINO~\cite{ssl4eos12} & ResNet50 & 25M & \textcolor{black}{Distillation} & \mbox{Sentinel-2} & 10m & 1M & 1 \\ \addlinespace[2mm]
DeCUR~\cite{decur} & ResNet50 & 25M & \makecell[t]{Contrastive\\learning} & \mbox{Sentinel-2} & 10m & 1M & 1 \\ \addlinespace[2mm]
ScaleMAE~\cite{scalemae} & ViT & 300M & MAE & RGB & 0.1-30m & 360k & 1 \\ \addlinespace[3mm]
DOFA~\cite{DOFA} & ViT & 300M & MAE & \begin{tabular}[c]{@{}c@{}}NAIP\\ \mbox{Sentinel-2} \\ \mbox{Sentinel-1} \\ EnMap \\ Gaofen \\ Landsat\end{tabular} & 1-30m & 8M & 1 \\ \addlinespace[3mm]
Satlas~\cite{satlas} & Swin & 100M & Supervised & \mbox{Sentinel-2} & 10m & NA & 1 \\ \addlinespace[2mm]
\mbox{Prithvi-EO-1.0~\cite{prithvi_v1}} & ViT & 100M & MAE & HLS & 30m & 250k & 3 \\ \addlinespace[2mm]
\mbox{Prithvi-EO-2.0} & ViT & 300M, 600M & MAE & HLS & 30m & 4.2M & 4 \\
\bottomrule
\end{tabularx}
\end{table*}


\begin{table*}[htbp]
\centering
\caption{Downstream tasks characteristics. T: Number of timestamps.}
\label{tab:downstreamtasks}
\small
\begin{tabularx}{1.8\columnwidth}{@{}Xc@{\hspace{25pt}}c@{\hspace{25pt}}c@{\hspace{25pt}}c@{}}
\toprule
\textbf{Domain} & \textbf{Task} & \textbf{Dataset size} & \textbf{T} & \textbf{Sensors} \\
\midrule
\multirow{4}{*}{{Disaster response}} & Flood detection & 446 (448 $\times$ 448)& 1 & Sentinel-2 \\ 	
& Wildfire scar detection & 805 (448 $\times$ 448)& 1 & HLS \\ 
& Burn scar intensity & 5,692 (\textcolor{black}{224} $\times$ \textcolor{black}{224}) & \textcolor{black}{3} & HLS \\
& Landslide detection & 3,799 (224 $\times$ 224)& 1 & Sentinel-2 \\
\midrule
\multirow{3}{*}{Land cover and crop} & Segmentation (US) & 5,000 (224 $\times$ 224) & 3 & HLS \\
& Classification (EU) & 335,125 (64 $\times$ 64) & up-to-12 & Sentinel-2 \\
& \textcolor{black}{Segmentation (France)} & \textcolor{black}{2,433 (128 $\times$ 128)} & \textcolor{black}{38 to 61} & \textcolor{black}{Sentinel-2} \\
\midrule
\multirow{2}{*}{Ecosystem dynamics} & Gross primary product & 975 (50 $\times$ 50) & 1 & HLS + MERRA2 \\ 
& Above Ground Biomass & 3,400 (256 $\times$ 256) & \textcolor{black}{12} & Sentinel-2 + Sentinel-1 \\ 
\bottomrule
\end{tabularx}
\end{table*}

\subsubsection{Disaster response}

\paragraph{Flood Mapping}\label{sec:flood_mapping_desc}

Sen1Floods11~\cite{bonafilia2020sen1floods11} is a surface water \textcolor{black}{dataset} including labels for water and land, and related Sentinel-1 and Sentinel-2 imagery. The dataset consists of 446 \textcolor{black}{512~$\times$~512} images covering 14 biomes, 357 ecoregions, and 6 continents of the world across 11 flood events between 2018 and 2020. The dataset and a reference training and evaluation code are available on GitHub\footnote{\url{https://github.com/cloudtostreet/Sen1Floods11}}. This dataset was used for testing Prithvi-EO-1.0, where the model showed beyond-state-of-the-art performance~\cite{prithvi_v1}. Here, we \textcolor{black}{compare it} with the newer Prithvi-EO version.

The \textcolor{black}{512~$\times$~512} images were resized to \textcolor{black}{448~$\times$~448} as \textcolor{black}{the first} is not divisible by the patch size of the 600M models (\textcolor{black}{14~$\times$~14}). \textcolor{black}{This guarantees} that all models are trained and validated with the uncropped and unpadded images. To increase data variability, random vertical and horizontal flips were applied as part of the data augmentation process. The associated masks present two categories: 0 for land and 1 for water. For all models, we used an UPerNet~\cite{upernet} decoder and tuned hyperparameters with \textcolor{black}{TerraTorch-iterate} (\textcolor{black}{ten} trials, tuning weight decay and learning rate). Then, we repeated the training with the best validation hyperparameters \textcolor{black}{ten} times with different seeds and reported results on the test set. The loss used was cross-entropy and the models ran for 50 epochs with early-stopping patience set to 20 epochs. We used the same six bands as in the pretraining set.

\textcolor{black}{For this task, we report the mIoU and the IoU for the water class. Additionally, we provide the \textcolor{black}{micro-averaged} F1 score.}

\paragraph{Wildfire Scar Mapping}

The wildfire scars dataset contains shapefiles of both intentional and wildfires burning from 2018--2021 as described in~\cite{prithvi_v1}.
This dataset\footnote{\url{https://huggingface.co/datasets/ibm-nasa-geospatial/hls_burn_scars}} was \textcolor{black}{also} used for testing Prithvi-EO-1.0, where the model showed state-of-the-art performance~\cite{prithvi_v1}. \textcolor{black}{Therefore, we compare it with} the newer Prithvi-EO version.

The \textcolor{black}{512~$\times$~512} images were \textcolor{black}{also} resized to \textcolor{black}{448~$\times$~448}, as explained in \textcolor{black}{section~\ref{sec:flood_mapping_desc}}. \textcolor{black}{For data augmentation, we used} random vertical and horizontal flips along with random crop to \textcolor{black}{224~$\times$~224}. The corresponding masks \textcolor{black}{show} 0 for unburnt areas and 1 for burnt areas. \textcolor{black}{We used} an UPerNet decoder~\cite{upernet} and ran hyperparameter tuning with \textcolor{black}{TerraTorch-iterate} \textcolor{black}{for all models} (\textcolor{black}{ten} trials, tuning weight decay and learning rate). The training with the best validation hyperparameters \textcolor{black}{was repeated} \textcolor{black}{ten} times with different seeds, and \textcolor{black}{the} results \textcolor{black}{are reported} on the test set. \textcolor{black}{We used} cross-entropy \textcolor{black}{as loss} and \textcolor{black}{set maximum epochs to} 50, with early-stopping patience \textcolor{black}{of} 20 epochs.

\textcolor{black}{In this case, we report the mIoU, the IoU for the burn scar class and the aggregated F1 score.}

\paragraph{Burn Intensity Mapping}

We collected burn scar intensity data using paired HLS imagery and burn scar intensity classifications from Burned Area Emergency Response (BAER)~\cite{napper2006burned}, which evaluates fire impacts on vegetation, soil, and watershed functionality. BAER teams rapidly analyze post-fire conditions to stabilize affected landscapes and mitigate hazards.

The dataset includes 5,692 \textcolor{black}{images} capturing spatial burn scar data and top-of-atmosphere (TOA) reflectance values across three temporal stages: pre-burn, during-burn, and post-burn\textcolor{black}{. The data was collected} from 2018 to 2023, over the western CONUS (126$^o$\textcolor{black}{W} to -104$^o$\textcolor{black}{W}\textcolor{black}{,} 32.5$^o$\textcolor{black}{N} to 48$^o$\textcolor{black}{N}). Burn scar intensity classifications, ranging from 0 (no burn) to 4 (high severity), were derived from burn severity analysis. The HLS imagery was carefully selected to match the temporal criteria for pre-fire and post-fire stages, ensuring that changes in burn intensity and reflectance values were accurately captured. Each fire event is represented by paired 224~$\times$~224 \textcolor{black}{images}: one for burn scar intensity and another for TOA reflectance across six spectral bands (Blue, Green, Red, NIR, SWIR\textcolor{black}{1}, SWIR\textcolor{black}{2}). The images were visually inspected to confirm burn scar visibility and cloud-free conditions. The dataset was further refined \textcolor{black}{by} excluding entries with more than 25\% missing or zero values, ensuring high-quality data for training \textcolor{black}{AI} models. 
The dataset's design ensures reproducibility and usability for wildfire monitoring and environmental research.

\begin{figure}[htp]
\centering
\includegraphics[width=0.9\columnwidth]{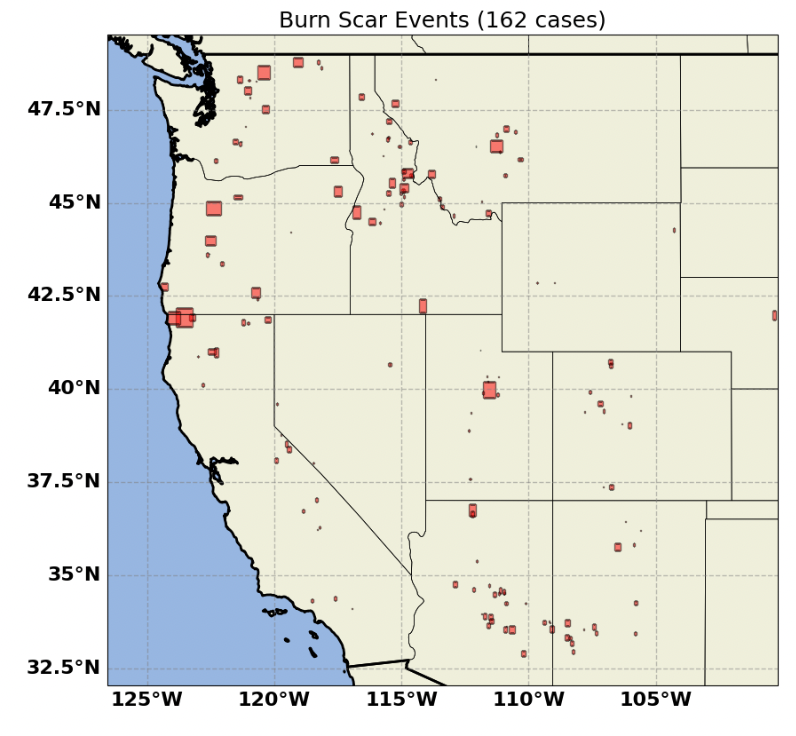}
\caption{Locations of the burn intensity dataset used in training and testing.}
\label{fig:burn_intensity_map}
\end{figure}

Out of \textcolor{black}{the} 5,692 \textcolor{black}{224~$\times$~224} images, 80\% \textcolor{black}{were} used for training and 20\% for \textcolor{black}{testing.}
The architecture for this task consisted of the Prithvi encoder along with a\textcolor{black}{n UperNet} decoder. 
The model was trained for \textcolor{black}{50} epochs \textcolor{black}{using a parameter-efficient fine-tuning technique called Low-Rank Adaptation, or LoRA~\cite{lora}}, with weighted cross-entropy as the loss function. \textcolor{black}{We set the learning rates to 5e-4 (encoder) and 5e-5 (decoder), with a cosine annealing schedule, weight decay of 0.05 and random horizontal flips for data augmentation.} \textcolor{black}{As baseline for comparison, we used a regular U-Net~\cite{unet} and a larger U-Net version with a ResNet-50 backbone. For this use case, we provide the mIoU along with per class IoU.}

\paragraph{Landslide Detection}

The Landslide4Sense (L4S) dataset~\cite{L4S-2022} is a comprehensive benchmark for large-scale landslide detection research, utilizing multi-source satellite imagery to study landslide-affected areas globally from 2015 to 2021. It includes 14 data layers: Sentinel-2 multispectral bands (B1–B12), a digital elevation model (DEM), and slope data from ALOS PALSAR~\cite{L4S-2022}. All layers are resized to a resolution of 10~m per pixel and labeled at the pixel level as \textit{landslide} or \textit{non-landslide}; \textcolor{black}{the images have 128~$\times$~128 pixels of size}. The dataset is divided into training, validation, and test sets, covering diverse mountainous regions worldwide. The training data comes from four sites: Iburi-Tobu (Hokkaido), Kodagu (Karnataka), Rasuwa (Bagmati), and western Taitung County, comprising 3,799 image patches. The validation and test sets include 245 and 800 \textcolor{black}{images}, respectively.

To fine-tune the Prithvi-EO-2.0 models, we incorporated a decoder consisting of four deconvolutional layers, along with layer normalization and activation functions, to map the model outputs from embeddings back to the original input image size. \textcolor{black}{A final set of two convolutional layers }
was then applied to project the latent space to class probabilities. 
For the Prithvi-EO-2.0-600M configuration, which uses a patch size of 14~$\times$~14, this decoder architecture does not allow for the perfect restoration of embeddings to the input image size. To address this issue, we added a bilinear interpolation layer following the \textcolor{black}{final convolutional layers} to resize the output and ensure it matches the original input image dimensions.

The models were fine-tuned using the data splits described \textcolor{black}{above}. Six-band HLS reflectance images were selected and upsampled \textcolor{black}{to} 224~$\times$~224 to match the configuration of \textcolor{black}{the} Prithvi\textcolor{black}{-EO-2.0} backbone. Data augmentation was applied by randomly flipping images both vertically and horizontally. \textcolor{black}{We considered} two different loss functions: weighted cross-entropy (wCE) loss, with a weighted ratio of 2:8 for negative and positive pixels, and Lovasz loss~\cite{berman2018lovasz}. Gradient descent was performed using the Adam optimizer with a learning rate of 1e-5. Additionally, a cosine learning rate scheduler with warm-up and restart~\cite{loshchilov2016sgdr} was utilized to enhance model convergence. The model\textcolor{black}{s were} trained for 100 epochs with a batch size of 8, and the checkpoint\textcolor{black}{s} with the lowest validation loss \textcolor{black}{were} selected for evaluation.

\textcolor{black}{For this task,} U-Net~\cite{unet} and U-Net++~\cite{zhou2018unet++} \textcolor{black}{were used as baselines models for comparison, both with a ResNet50 as the backbone}.
These are popular and powerful models that have been used to support a wide range of environmental mapping tasks, including landslide mapping~\cite{li2023assessment, L4S-2022,ling2024optimized}. It is worth noting that this dataset was also used in the 2022 Landslide4Sense competition, with several winning models reported by~\cite{L4S-2022}. However, those results were based on customized training strategies, such as self-training using test data, pseudo-label refinement, and ensemble methods. In contrast, our study focuses on evaluating the performance of \textcolor{black}{the Prithvi-EO-2.0 models} on the L4S task and comparing it with existing deep learning models under a standardized training protocol (\textcolor{black}{i.e.}, consistent data splits, augmentation techniques, and training platforms) to ensure a fair comparison.

\textcolor{black}{We evaluated model performance under two fine-tuning data regimes: using the full training set and a reduced subset comprising approximately 1\% of the full data. The latter allows us to assess the models' few-shot learning capabilities.}
\textcolor{black}{Three evaluation} metrics \textcolor{black}{were considered}: mIoU, F1 score, and \textcolor{black}{micro-averaged} accuracy (\textcolor{black}{m}Acc).

\subsubsection{Land cover and crop mapping}

\paragraph{Multi-Temporal Crop Segmentation in the United States}

\textcolor{black}{The crop segmentation task was one of the datasets used to evaluate} Prithvi-EO-1.0, where the model showed state-of-the-art performance~\cite{prithvi_v1, hls-multi-temporal-crop-classification}. \textcolor{black}{In this work, we compare it with} the newer Prithvi-EO version. \textcolor{black}{We also include a U-Net as an additional baseline model for this task.}

The \textcolor{black}{224~$\times$~224} images were used for both training and validation without cropping. To enhance data variability, random vertical and horizontal flips were applied as part of the data augmentation process. The corresponding masks contain thirteen classes: 1) "Natural Vegetation", 2) "Forest", 3) "Corn", 4) "Soybeans", 5) "Wetlands", 6) "Developed/Barren", 7) "Open Water", 8) "Winter Wheat", 9) "Alfalfa", 10) "Fallow/Idle Cropland", 11) "Cotton", 12) "Sorghum", and 13) "Other".

\textcolor{black}{For fine-tuning, we used the weighted cross-entropy as the loss function. All Prithvi-EO encoders} were combined with a simple segmentation decoder consisting of convolutional and upsampling blocks\textcolor{black}{, and trained for 80 epochs. Besides the mIoU, we also provide the aggregated accuracy for all models.}

\paragraph{Multi-Temporal Land Cover and Crop Classification in Europe}

Sen4Map~\cite{sharma2024} is a benchmark dataset designed for multi-temporal land-cover and crop classification tasks with Sentinel-2 satellite data. Sen4Map comprises 335,125 time series of 64~$\times$~64 pixels multispectral \textcolor{black}{images}, extracted from Sentinel-2 tiles and annotated with geolocated land-cover data collected through extensive in situ surveys using LUCAS points~\cite{dandrimont2020}. The dataset includes time series of varying lengths from 2018, covering countries across the European Union. We utilized the monthly composites, feeding the models with time series of length 12, in line with the original \textcolor{black}{approach~\cite{sharma2024}}. The spatial resolution of the images in Sen4Map is natively 10~m or upsampled to 10~m from the native 20~m resolution, depending on the \textcolor{black}{band}. Each image covers an area of 640~m~$\times$~640~m, which contrasts with our pretraining dataset, where the images are 224~$\times$~224 pixels with a resolution of 30~m, covering approximately 6,720~m~$\times$~6,720~m—over 100 times the area of the Sen4Map samples. 

The patch-processing methodology further compounds the fine-tuning challenge. In Sen4Map, the original 64~$\times$~64-pixel \textcolor{black}{images} are cropped to smaller 15~$\times$~15 regions to focus on the central pixels, around the available label from LUCAS. This ensures that classification is based solely on the area of interest\textcolor{black}{,} minimizing the influence of irrelevant spatial context.
We adopted a similar setup to maintain a fair comparison. 
\textcolor{black}{However, 15~$\times$~15 pixels is comparable to the Prithvi-EO patch sizes, which means that the input image would have a single patch. Hence, to avoid changing the model's patch size and train the patch-embedding layer from scratch,} the 15~$\times$~15 patches were upscaled to 224~$\times$~224 pixels.

In addition to differences in spatial resolution and \textcolor{black}{image} size, the datasets vary in the spectral bands. The pretraining dataset includes six \textcolor{black}{bands from} HLS data, while the Sen4Map dataset provides four additional bands (NIR-Broad, Red-Edge 1, 2, and 3). 
\textcolor{black}{The} differences in spatial, spectral, and contextual characteristics \textcolor{black}{can imply a greater degree of adaptation for the Prithvi models, potentially extending or complicating the fine-tuning process.}

We fine-tuned the original Prithvi-EO-1.0-100M model, Prithvi-EO-2.0-300M, and Prithvi-EO-2.0-600M, on different \textcolor{black}{ratios} of the Sen4Map training data. We \textcolor{black}{used the weighted F1 score to compare} them with the baseline ViViT~\textcolor{black}{\cite{vivit} adopted} in Sen4Map, which was recreated to be trained from scratch on the \textcolor{black}{data subsets}.

\paragraph{\textcolor{black}{Multi-Temporal Crop Segmentation with PASTIS}}

\textcolor{black}{The PASTIS dataset contains 2,433 sequences of 128~$\times$~128 multispectral images, from four regions in France. Sequence lengths range from 38 to 61 frames, collected at irregular intervals between September 2018 and November 2019. The samples include ten channels from Sentinel-2's non-atmospheric bands, resampled to a 10~m resolution. The dataset provides segmentation labels for 18 crop types plus a background class and it is highly imbalanced: the most frequent classes appear over 50 times more often than the rarest. The authors estimate that about 28\% of the images have some cloud cover~\cite{pastis}.}

\textcolor{black}{Given the large and variable sequence lengths, we standardized them to 20 frames per sample. The sequences were divided into 20 equal-width temporal bins, and one image was randomly selected from each bin. This strategy ensures coverage across the entire sequence, better preserving temporal diversity, which could be compromised with simple truncation or random sampling. The frame selection was fixed for validation and test. Additionally, we removed samples with high cloud coverage as it was beneficial for all tested models.}

\textcolor{black}{We compared Prithvi-EO-2.0 with the PASTIS baseline, U-TAE~\cite{pastis} (a U-Net architecture enhanced with temporal attention mechanisms), and also with other foundation models: Satlas~\cite{satlas}, DOFA~\cite{DOFA}, and Presto~\cite{presto}.}

\textcolor{black}{All models were fine-tuned with the same procedure. Data augmentation included random resized cropping, horizontal flipping, and rotation. The dataset followed the original split~\cite{pastis}: folds 1–3 for training, 4 for validation, and 5 for testing. Each GFM was paired with an UPerNet decoder~\cite{upernet}, and fine-tuned with 100\% and 10\% of the training data.}

\textcolor{black}{Since Satlas and DOFA do not process image sequences, frames were passed independently through their backbones, and the features were concatenated. Except for Presto, the GFM encoders were connected to a Lightweight Temporal Attention Encoder (LTAE)~\cite{ltae} followed by a Temporal Aggregator~\cite{ltae} before the UPerNet decoder. With long input sequences, these models generate a large number of tokens, and directly connecting them to the UPerNet would have resulted in a substantial increase of decoder parameters, hindering training and efficiency.}

\textcolor{black}{The GFMs were trained using LoRA~\cite{lora}, cross-entropy loss, and the AdamW optimizer, with different learning rates for the encoder and decoder: 5e-4 and 5e-5, respectively.
We set the weight decay to 0.05 and trained the models for 50 epochs with early stopping (patience of 10 epochs). The mIoU was used as the evaluation metric.}

\subsubsection{Ecosystem dynamics}
\paragraph{Above Ground Biomass Estimation}

Above Ground Biomass (AGB) is a crucial climate variable that encompasses all living biomass above the soil, essential for understanding land cover/use changes and carbon sinks. Although direct measurement of AGB is challenging, remote sensing methods and machine learning techniques can be used to estimate it, providing comprehensive spatial and temporal data. The BioMassters dataset was generated to explore deep learning approaches for predicting yearly \textcolor{black}{AGB} in Finnish forests using multimodal, multi-temporal Sentinel-1 and Sentinel-2 satellite data from 2016 to 2022~\cite{nascetti2023biomassters}. The dataset comprises 11,462 AGB reference images, each paired with 12 months of corresponding satellite imagery, covering 2,560~$\times$~2,560-meter forest patches, with AGB measured in metric tonnes per pixel and derived from LiDAR data calibrated with in-situ measurements. The winning model from the BioMassters data challenge was used as the \textcolor{black}{baseline} for comparison with \textcolor{black}{Prithvi-EO-2.0}. This model is based on a U-Net architecture with temporal attention encoding~\cite{garnotPanopticSegmentationSatellite2021}, using all 12 \textcolor{black}{timestamps} and all 15 \textcolor{black}{bands} from Sentinel-1 and Sentinel-2 imagery as inputs for each yearly AGB prediction. \textcolor{black}{It} achieved an RMSE of 27.49 \textcolor{black}{in} the holdout test set. 

The dataset was divided into 80\% training and 20\% validation sets for consistent experimental use. \textcolor{black}{Also, we tested different input configurations given the differences between the BioMassters dataset and Prithvi-EO pretraining data: the first includes 12 observations of Sentinel-1 and -2 data for each target AGB image, while the second considers four frames from HLS data. The configurations include:}
\begin{itemize}
    \item \textcolor{black}{6-band Sentinel-2, 4 frames}
    \item \textcolor{black}{6-band Sentinel-2, Sentinel-1, 4 frames}
    \item \textcolor{black}{6-band Sentinel-2, 12 frames}
    \item \textcolor{black}{6-band Sentinel-2, Sentinel-1, 12 frames}
    \item \textcolor{black}{11-band Sentinel-2, Sentinel-1, 12 frames}
\end{itemize}
\textcolor{black}{where the six bands are the same as used in the HLS pretraining data and the four timestamps} were chosen out of 12 based on the quality of the Sentinel-2 data, prioritizing scenes with no missing data, followed by scenes with the lowest cloud cover. \textcolor{black}{The highest performing configuration was then selected as the subject of subset training, where performance of the model was tested when trained on subsets of 50\%, 20\%, 10\%, and 5\% of the original training set}.


\textcolor{black}{We fine-tuned the Prithvi-EO-2.0-300M encoder paired with a FCN decoder and a regression head. We used the AdamW optimizer, with a learning rate of 6e-5 and weight decay of 0.05. As data augmentation, random flips and rotations were applied. For the cases using the 6-band Sentinel-2 data, we experimented with the efficient fine-tuning technique LoRA~\cite{lora}. In the experiments utilizing Sentinel-1 data, we included the ascending and descending VV, VH, and RVI bands.}



\paragraph{Estimation of Gross Primary Productivity (GPP) at Globally Distributed Sites}

Photosynthesis (Gross Primary Productivity (GPP)), the critical process converting solar radiation into energy and materials that sustain life, represents the largest global carbon cycle flux, yet remains challenging to quantify accurately without direct observation~\cite{keenan2023constraint}. Data-driven approaches using machine learning to integrate satellite data with site-level information (e.g.\textcolor{black}{,} from eddy covariance measurement networks~\cite{baldocchi2020eddy}) provide diagnostic constraints on global GPP, complementing process-based models~\cite{jung2020scaling}. However, these approaches face challenges due to scale mismatches between in-situ measurements and satellite data, as well as generalization issues, which could potentially be mitigated by using foundation models like Prithvi\textcolor{black}{-EO} that embed generalized representations of high-resolution data~\cite{kang2025cedar}.

This downstream application aims to evaluate the potential of Prithvi\textcolor{black}{-EO-2.0} to estimate GPP at globally distributed eddy covariance sites. We utilize\textcolor{black}{d} daily GPP data from 37 eddy covariance flux towers, obtained from FLUXNET~\cite{pastorello2020fluxnet2015}, AmeriFlux\footnote{\url{https://ameriflux.lbl.gov/data/flux-data-products/}}, and ICOS~\cite{icos_flux} networks. We use datasets processed by the ONEFLUX pipeline for high-quality and consistent data. The target variable for analysis is GPP estimate derived from Net Ecosystem Exchange (NEE) measurements using the nighttime approach (\textit{GPP\_NT\_VUT\_REF}). We selected daily GPP data with at least 60\% of high-quality hourly or half-hourly data for temporal aggregation. \textcolor{black}{The latitude and longitude of the flux tower sites were recorded to retrieve HLS and MERRA-2 data from corresponding locations. For HLS, the recorded coordinates define the center of the images, while for MERRA-2 variables, the nearest grid point is used}.

The dataset combines three primary components: 1) HLS \textcolor{black}{six}-band reflectance data organized as 50~$\times$~50~$\times$~6 pixels (TIFF files), 2) 10-dimensional Modern-Era Retrospective analysis for Research and Applications version 2 (MERRA-2) atmospheric and land surface variables capturing environmental conditions, and 3) daily GPP measurements derived using the night-time partitioning approach. HLS and MERRA-2 data are centered on flux tower locations. The GPP data span 37 globally distributed flux tower sites from 2018 to 2021, encompassing 975 samples. MERRA-2 variables include minimum, maximum, and mean temperatures, soil moisture, heat flux, radiation, precipitation, and other key environmental metrics, identified using \textcolor{black}{the codes} M2T1NXSLV and M2T1NXLND.

The dataset preparation process involves meticulous filtering and quality checks. HLS data are pre-selected based on a 25\% maximum cloud threshold and 75\% minimum spatial coverage threshold, with further filtering to remove scenes with $>$2\% snow cover and $>$5\% cloud cover. HLS reflectance values are scaled accordingly. MERRA-2 data are aggregated as daily mean values for each flux site. Instances with GPP values $\leq -\text{0.1 gCm}^{-\text{2}}\text{s}^{-\text{1}}$, indicative of poor data, are removed. Corresponding HLS and MERRA-2 inputs for each GPP record are matched, and additional filters exclude data with spatial mean Enhanced Vegetation Index (EVI) values $\leq -\text{0.1}$ across images, which signal snow or cloud contamination. This dataset provides a high-quality, integrated framework of remote sensing and ground observations for studying GPP dynamics under diverse environmental conditions.

\begin{figure}[htbp] 
\centering 
\includegraphics[width=\columnwidth]{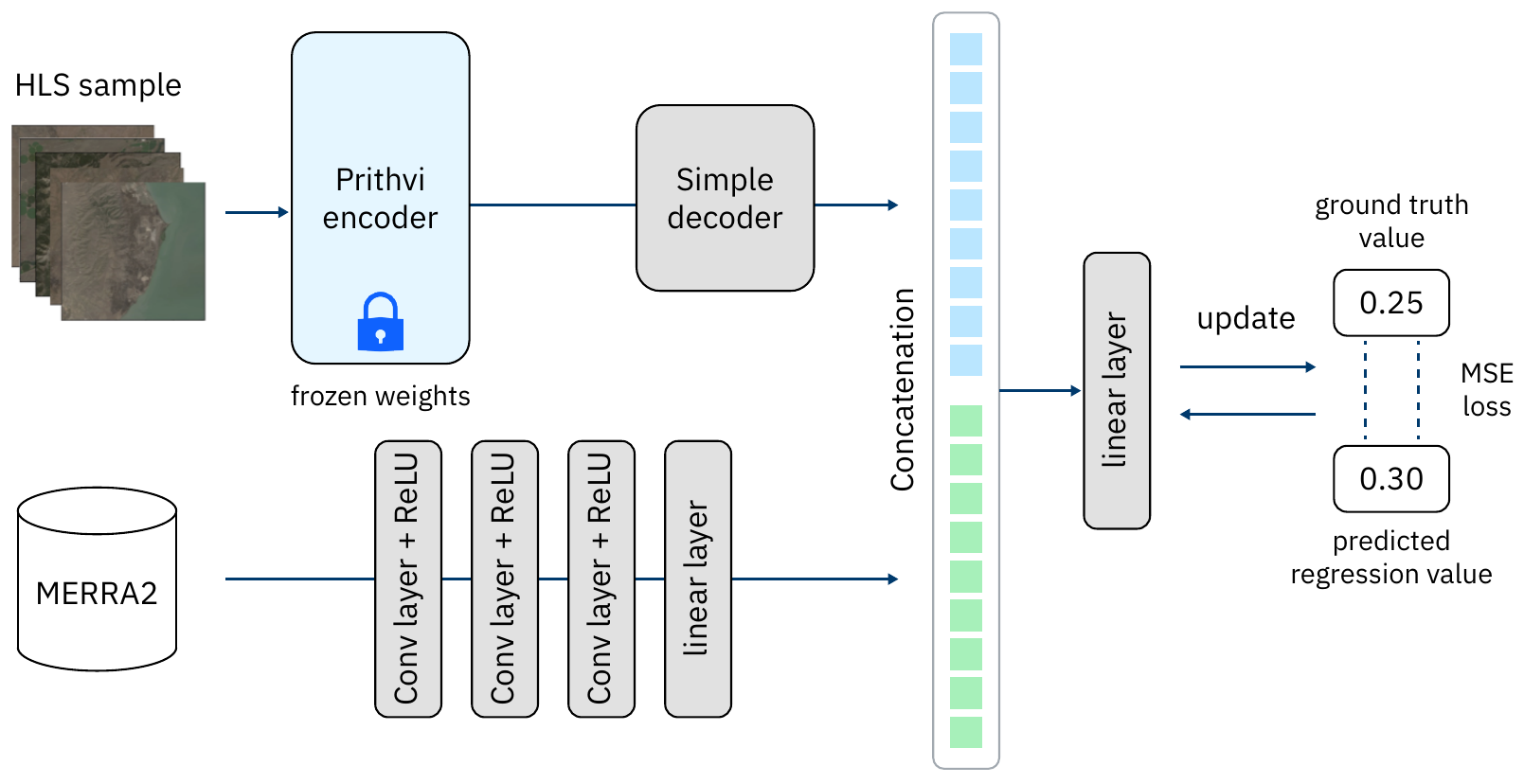} 
\caption{Workflow to fine-tune Prithvi\textcolor{black}{-EO-2.0} for predicting GPP at flux observations sites. The \textcolor{black}{model} consists of two branches: the (frozen) Prithvi encoder \textcolor{black}{generating} HLS embeddings \textcolor{black}{later processed} by a simple decoder, and convolutional \textcolor{black}{layers} transforming MERRA-2 atmospheric and land surface variables. The decoder and convolved MERRA-2 outputs are concatenated and \textcolor{black}{processed with a linear layer} to predict GPP.} 
\label{fig:flux_model} 
\end{figure}

\textcolor{black}{The fine-tuning architecture developed for this task (Figure~\ref{fig:flux_model}) combines the Prithvi-\textcolor{black}{EO-2.0} encoder, which receives} the six-band HLS reflectance data, \textcolor{black}{with a simple convolutional network that processes the} ten input features from MERRA-2. \textcolor{black}{This approach is designed to facilitate extending the Prithvi\textcolor{black}{-EO-2.0} HLS representation with additional environmental variables to predict GPP values at flux tower sites.}
\textcolor{black}{More specifically}, the multispectral HLS input is passed through the \textcolor{black}{pre}trained and frozen Prithvi\textcolor{black}{-EO-2.0} encoder to \textcolor{black}{generate latent} HLS representations (Prithvi embeddings) of the flux observation sites\textcolor{black}{. These embeddings} are then passed through a \textcolor{black}{lightweight} decoder \textcolor{black}{composed of} two linear layers with ReLU activation. The MERRA-2 input features
are transformed in parallel via a separate branch consisting of three 2D convolutional \textcolor{black}{layers} with ReLU activations \textcolor{black}{followed by a} linear layer. The \textcolor{black}{Prithvi embeddings} and the MERRA-2 \textcolor{black}{latent} representation are both flattened, then concatenated and regressed to predict the GPP values using a linear layer. The models are trained by minimizing the mean squared error loss between the predicted GPP and \textcolor{black}{the} true GPP. 
Data splitting for training and evaluation follows a leave-one-year-out cross-validation approach, where three years are used for training and one for testing to \textcolor{black}{reduce} the challenges posed by the relatively small dataset size.

\textcolor{black}{We experimented} with all variants of Prithvi\textcolor{black}{-EO-2.0,} and quantify the models' skill in predicting GPP using the coefficient of determination, $R^2$ measure\textcolor{black}{,} between the predicted and \textcolor{black}{the} true GPP values. 
Prithvi\textcolor{black}{-EO-2.0} is compared \textcolor{black}{against three} baseline models based on random forest, XGBoost\textcolor{black}{, and ResNet18. The random forest and XGBoost models use} the spatial average of HLS bands across each \textcolor{black}{image} and the MERRA-2 data, representing the state-of-the-art approaches~\cite{kang2025cedar}. \textcolor{black}{These models} also incorporate vegetation indices, including EVI, Normalized Difference Vegetation Index (NDVI), Normalized Difference Water Index (NDWI), Green Chlorophyll Index (GCI), near-infrared reflectance of vegetation (NIRv), and kernel NDVI. \textcolor{black}{The ResNet model uses the same inputs as Prithvi-EO-2.0.}

\section{Results and Discussion}

\textcolor{black}{In this section, we provide and discuss the results of the evaluation experiments. We begin with the benchmarking results followed by the three categories of downstream tasks.}

\subsection{\textcolor{black}{Benchmarking}}

\begin{figure}[htbp]
     \begin{subfigure}{\columnwidth}
         \centering
         \includegraphics[width=\textwidth]{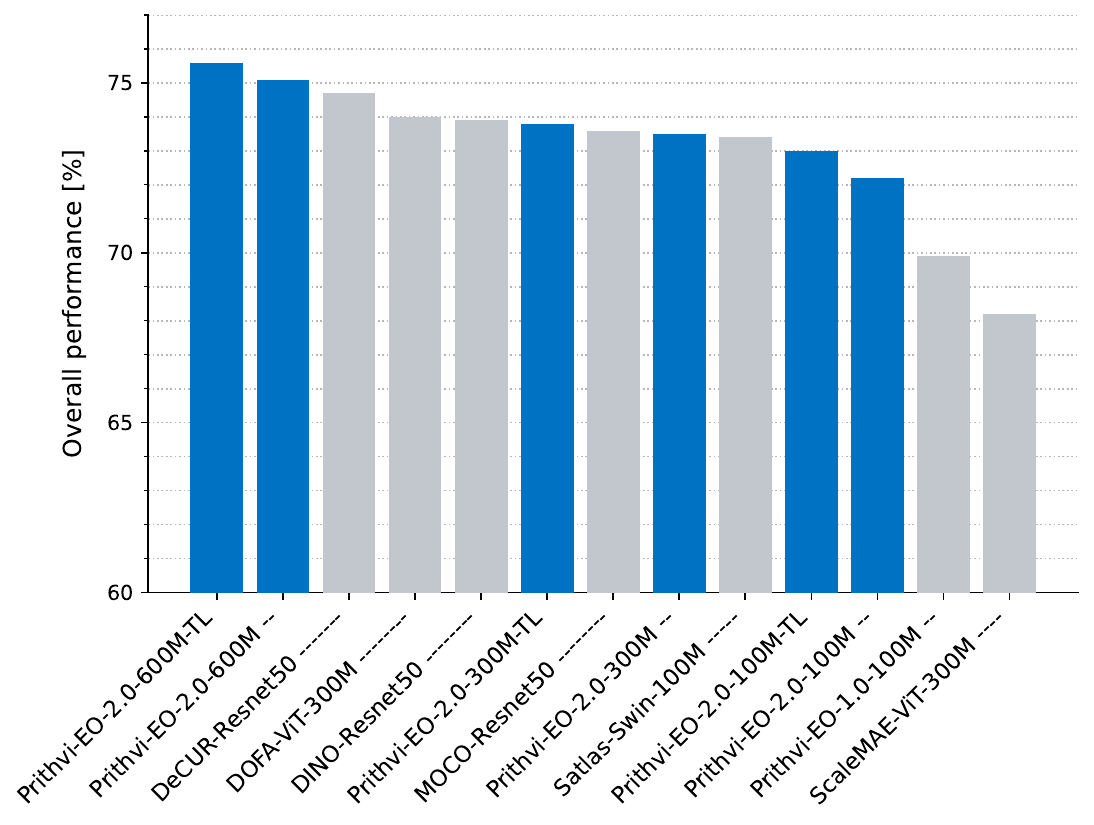}
         \caption{}
     \end{subfigure}
     \hfill
     \begin{subfigure}[b]{\columnwidth}
         \centering
         \includegraphics[width=\textwidth]{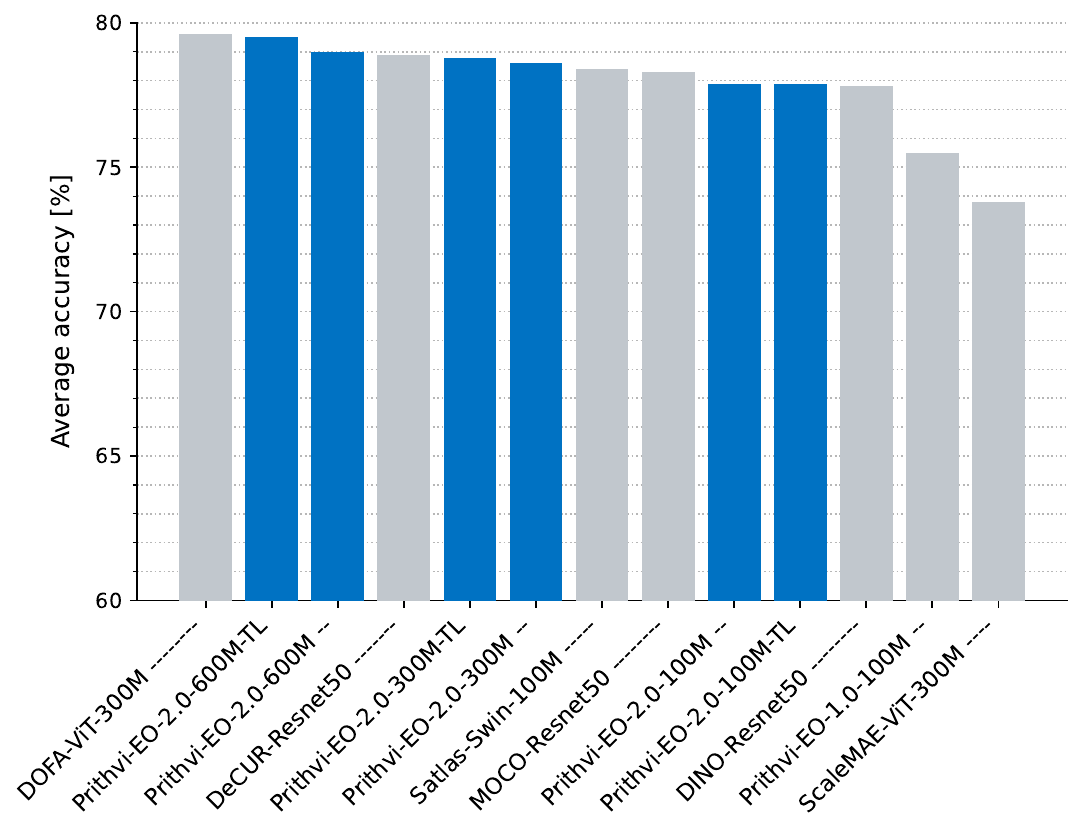}
         \caption{}
     \end{subfigure}
     \hfill
     \begin{subfigure}[b]{\columnwidth}
         \centering
         \includegraphics[width=\textwidth]{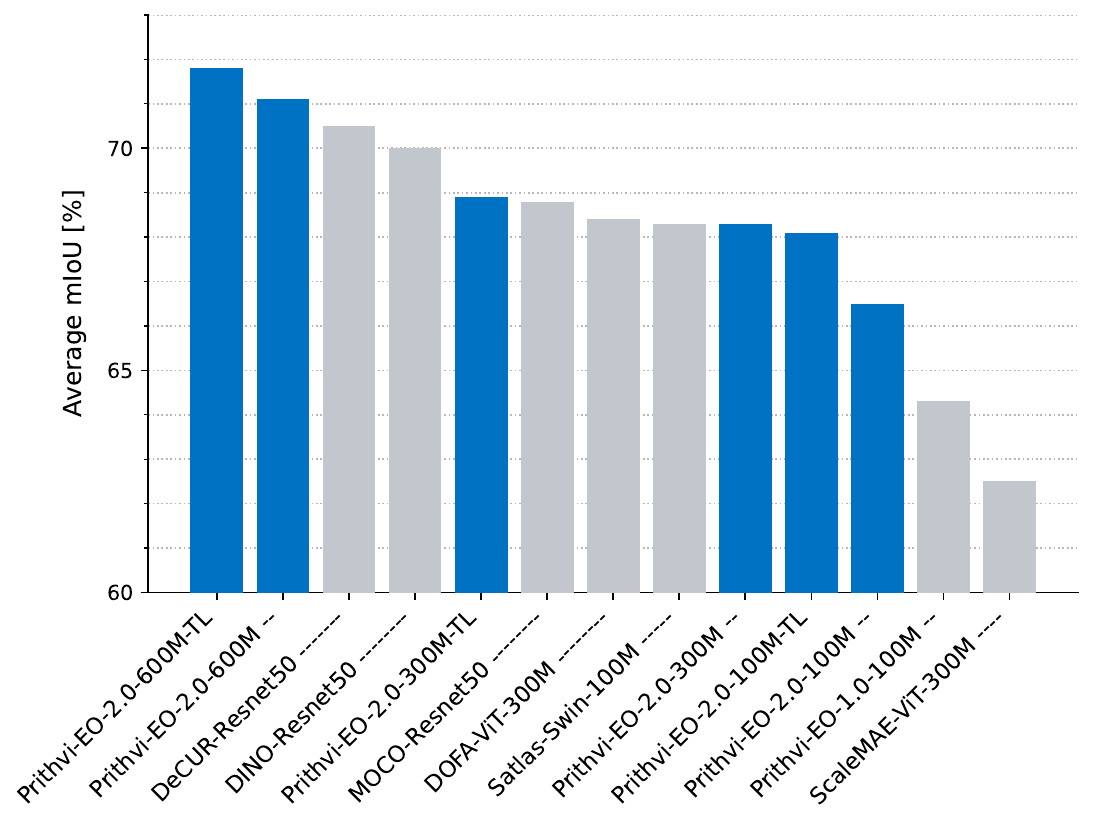}
         \caption{}
     \end{subfigure}
\caption{Aggregated performance across (a) all GEO-Bench datasets, (b) all classification tasks, and (c) all segmentation tasks. The Prithvi-EO-2.0 models are highlighted in blue.}
\label{fig:averaged_perf_geobench}
\end{figure}

\textcolor{black}{Figure~\ref{fig:averaged_perf_geobench} shows the aggregated} results of benchmarking with GEO-Bench datasets, including the average performance across the classification and segmentation datasets, as well as the overall performance across all 12 datasets. The results indicate that for the classification tasks, DOFA~\cite{DOFA}, Prithvi-EO-2.0-600M-TL and Prithvi-EO-2.0-600M are the top performing models. For segmentation, Prithvi-EO-2.0-600M-TL and Prithvi-EO-2.0-600M show the highest performance, \textcolor{black}{with} a noticeable margin \textcolor{black}{over} the next best models, DeCUR~\cite{decur} and DINO~\cite{ssl4eos12} ResNets. Overall, Prithvi-EO-2.0-600M-TL and Prithvi-EO-2.0-600M obtained the best combined performance across all 12 datasets. \textcolor{black}{Also, a}ll versions of Prithvi-EO-2.0 performed better than their predecessor Prithvi-EO-1.0 highlighting the improvement of the new version of the model. 

\begin{figure*}[htp]
    \centering
    \begin{subfigure}{\textwidth}
    \includegraphics[width=\textwidth,keepaspectratio]{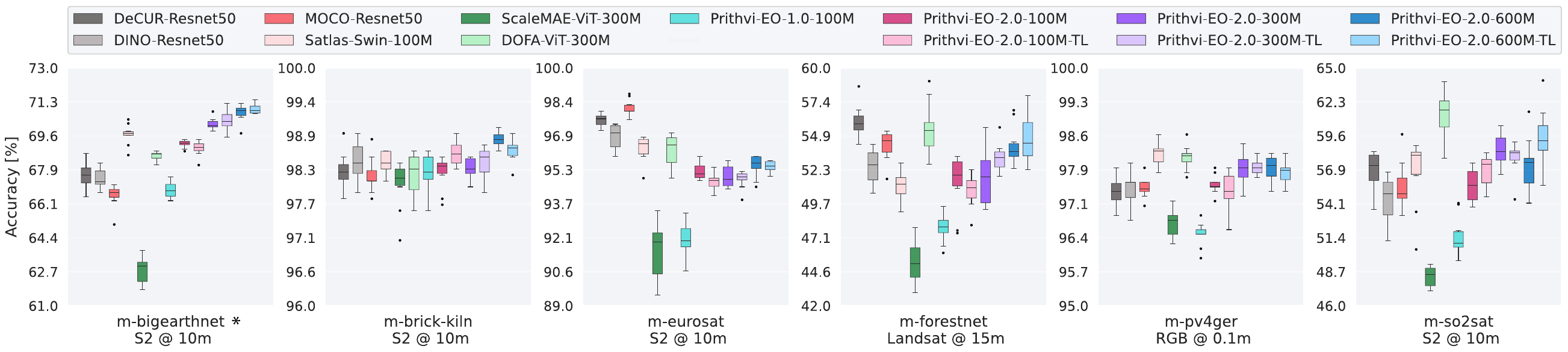}
    \caption{}
    \end{subfigure}
    
    \begin{subfigure}{\textwidth}
    \includegraphics[width=\textwidth]{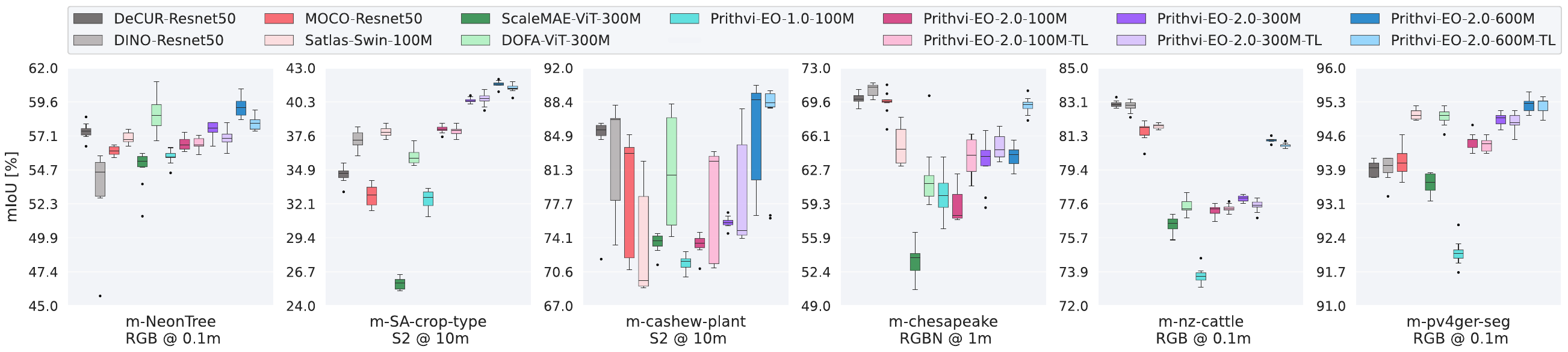}
    \caption{}
    \end{subfigure}
    
    \caption{\textcolor{black}{Distribution of performance across all models from 10 repeated runs for the GEO-Bench (a) classification and (b) segmentation tasks in terms of accuracy (F1 score for m-bigearthnet) and mIoU, respectively. We list input sensor and resolution for each dataset.}}
    \label{fig:class_seg_dist_geobench}    
\end{figure*}

Figure~\ref{fig:class_seg_dist_geobench} \textcolor{black}{provides} a more detailed view \textcolor{black}{of} the distribution of \textcolor{black}{the ten repeated experiments for each model} across \textcolor{black}{the} different \textcolor{black}{tasks}. For some datasets, multiple models perform equally well, with accuracies $\geq$~97\% (e.g., m-brik-klin or m-pv4ger). Among the remaining datasets, Prithvi-EO-2.0-600M-TL and Prithvi-EO-2.0-600M outperformed other models in four out of six tasks targeting medium resolution \textcolor{black}{imagery} (i.e.\textcolor{black}{, Sentinel-}2 at 10~m)\textcolor{black}{. In particular, tasks such as} m-bigearthnet and m-SA-crop-type \textcolor{black}{show} better \textcolor{black}{results from} all versions of Prithvi-EO-2.0-600M and Prithvi-EO-2.0-300M compared to the other models. This is \textcolor{black}{noteworthy} because these tasks \textcolor{black}{involve} earth observation applications like land use classification using the CORINE Land Cover product~\cite{corine}, heavily focused on forest and crop classification, which \textcolor{black}{are tasks associated with} strong seasonality. The strong performance \textcolor{black}{of} Prithvi-EO-2.0-600M-TL highlights the importance of designing \textcolor{black}{models} capable of learning relevant multi-temporal features useful for key remote sensing tasks.

\textcolor{black}{In Figure~\ref{fig:class_seg_dist_geobench}, we can also observe that several models show a wide range of scores on the m-cashew-plant dataset. This variation is related to the presence of seven classes in the training and validation sets, while the test set only contains six: class 0 is absent. When a model correctly avoids assigning any pixel of the test set as class 0, that class is excluded from the average IoU calculation. However, if a model does predict class 0, it becomes part of the average computation, which lowers the aggregated score.}

The performance comparison across models revealed four key insights. First, the benefits of using a larger, global dataset are evident after comparing the performance of Prithvi-EO-2.0-100M and Prithvi-EO-1.0-100M. Both models have identical architectures and model sizes, but the former was pretrained on the larger dataset and showed a 3\% improvement on the overall GEO-Bench score, with differences that are even larger for some of the tasks. Second, increasing the number of parameters correlated with better performance, evidenced in larger Prithvi-EO-2.0 models performing better than smaller models. Third, the models \textcolor{black}{incorporating} temporal and location embeddings had a higher overall performance compared to \textcolor{black}{their non-TL counterparts}, which is especially \textcolor{black}{interesting} considering that not all datasets included in GEO-Bench contain temporal and location information. Finally, Prithvi-EO-2.0 showed good performance on high-resolution tasks, despite being pretrained only with 30~m data. This improved performance on higher resolution tasks shows the versatility of our model for multi-resolution applications, including those that are more typical in computer-vision tasks, such as tree crown identification or cattle tracking using drone imagery.

\subsection{Disaster response}

\subsubsection{\textcolor{black}{Flood Mapping}}
Table~\ref{tab:sen1floods11} shows the results of our experiments on the test set \textcolor{black}{of the Sen1Floods11 dataset}. Prithvi-EO-2.0 improves over its previous version. Given the unbalanced nature of this dataset, where the land class is predominant and \textcolor{black}{easily identifiable}, improvements \textcolor{black}{in the average metrics are not substantial}. However, we \textcolor{black}{observe} a clear difference \textcolor{black}{when comparing the IoU for the water class between} Prithvi-EO-1.0-100M and \textcolor{black}{the new versions, with Prithvi-EO-2.0-600M-TL achieving an IoU 3.5 points higher than its predecessor.}

\begin{table}[htbp]
\centering
\caption{Results on the \textcolor{black}{Sen1Floods11 test set}. \textcolor{black}{TL: pretrained with temporal and location embeddings.}}
\label{tab:sen1floods11}
\small
\begin{tabular}{@{}lccc@{}}
\toprule
\textbf{Model} & \textbf{mIoU (std)} & \textbf{mF1 (std)} & \makecell{\textbf{IoU (std)} \\ \textbf{\textcolor{black}{water}}} \\
\midrule
Prithvi-EO-1.0-100M    & 88.3 (0.3) & 97.3 (0.1) & 79.6 (0.5) \\ 		
Prithvi-EO-2.0-300M    & 89.7 (0.3) & 97.6 (0.1) & 82.2 (0.6) \\ 		
Prithvi-EO-2.0-300M-TL & 90.0 (0.2) & 97.7 (0.1) & 82.6 (0.3) \\ 		
Prithvi-EO-2.0-600M    & 89.9 (0.6) & 97.6 (0.2) & 82.5 (1.0) \\ 		
Prithvi-EO-2.0-600M-TL & \textbf{90.3 (0.3)} & \textbf{97.7 (0.1)} & \textbf{83.1 (0.5)} \\ 		
\bottomrule
\end{tabular}
\end{table}

\subsubsection{\textcolor{black}{Wildfire Scar Mapping}}

Table~\ref{tab:firescars} \textcolor{black}{presents} the performance of the models on the wildfire scar mapping test set. \textcolor{black}{We notice s}imilar trends to \textcolor{black}{those observed in} the flood detection task, with the new versions of the model outperforming Prithvi-EO-1.0. \textcolor{black}{Once again, the} differences become more \textcolor{black}{evident} when focusing on the IoU for the wildfire scar class, where the \textcolor{black}{gap} between \textcolor{black}{the previous} version of the model and Prithvi-EO-2.0-600M-TL is 5.6 points, \textcolor{black}{indicating the improved capabilities of our new models}.
\begin{table}[htbp]
\centering
\caption{Results on the wildfire scar mapping \textcolor{black}{test set}. \textcolor{black}{TL: pretrained with temporal and location embeddings.}}
\label{tab:firescars}
\small
\begin{tabular}{@{}lccc@{}}
\toprule
\textbf{Model} & \textbf{mIoU (std)} & \textbf{mF1 (std)} & \makecell{\textbf{IoU (std)} \\ \textbf{burn scar}} \\
\midrule
Prithvi-EO-1.0-100M & 86.9 (0.8) & 97.2 (0.2) & 76.8 (1.4) \\
Prithvi-EO-2.0-300M & 88.6 (0.5) & 97.6 (0.1) & 79.9 (0.8) \\
Prithvi-EO-2.0-300M-TL & 89.3 (0.9) & 97.8 (0.3) & 81.1 (1.6) \\
Prithvi-EO-2.0-600M & \textbf{90.5 (0.6)} & \textbf{98.1 (0.2)} & \textbf{83.2 (1.1)} \\
Prithvi-EO-2.0-600M-TL & \textbf{90.5 (0.7)} & \textbf{98.1 (0.2)} & \textbf{83.2 (1.1)} \\
\bottomrule
\end{tabular}
\end{table}

\subsubsection{\textcolor{black}{Burn Intensity Mapping}}

\textcolor{black}{In} Table~\ref{tab:burn_intensity_training} \textcolor{black}{we present} the results of the burn intensity predictions \textcolor{black}{on the test set}. The results showed that \textcolor{black}{both Prithvi-EO-2.0 models outperform the U-Net versions} in terms of \textcolor{black}{mIoU}. \textcolor{black}{Interestingly, although the mIoU is generally low for all models, they seem capable of distinguishing unburnt from burnt areas, with IoU\textsubscript{0} above 67\%. Prithvi-EO-2.0-600M achieves the higher IoU for class 0 (75.9\%), representing a significant improvement over the U-Net models. However, all models seem to struggle to differentiate between the various severity classes, with the Prithvi models demonstrating better overall performance. We believe this might be linked to two data-related aspects. The first is that a number of samples shows a high coverage of smoke in the ``during'' image, introducing noise the models must learn to filter out. The second is related to the label masks: they show discontinuity and a general pattern where very small regions of higher severity classes appear within a larger, more continuous regions of lower severity classes. These characteristics can be challenging for models with larger patch sizes, such as the Prithvi models. The U-Nets, on the other hand, despite having access to pixel-level information and typically showing better capabilities in capturing more local patterns, were unable to accurately distinguish the burn-severity classes.}


\begin{table}[htbp]
\centering
\caption{\textcolor{black}{Results on the Burn intensity test set: mIoU and IoU for each class from 0 (no burn) to 4 (high severity).}}
\label{tab:burn_intensity_training}
\small
\begin{tabular}{@{}l@{\hspace{10pt}}c@{\hspace{10pt}}c@{\hspace{10pt}}c@{\hspace{10pt}}c@{\hspace{10pt}}c@{\hspace{10pt}}c@{}}
\toprule
\textbf{Model} & \textbf{mIoU} & \textbf{\textcolor{black}{IoU\textsubscript{0}}} & \textbf{\textcolor{black}{IoU\textsubscript{1}}} & \textbf{\textcolor{black}{IoU\textsubscript{2}}} & \textbf{\textcolor{black}{IoU\textsubscript{3}}} & \textbf{\textcolor{black}{IoU\textsubscript{4}}} \\
\midrule
U-Net                 & \textcolor{black}{26.3} & \textcolor{black}{67.5} & \textcolor{black}{10.0} & \textcolor{black}{19.9} & \textcolor{black}{23.7} & \textcolor{black}{10.6} \\
U-Net (ResNet-50)     & \textcolor{black}{28.8} & \textcolor{black}{68.1} & \textcolor{black}{18.2} & \textcolor{black}{21.7} & \textcolor{black}{27.8} & \textcolor{black}{8.05} \\
Prithvi-EO-2.0-300M & \textcolor{black}{\textbf{31.2}} & \textcolor{black}{73.3} & \textcolor{black}{\textbf{22.7}} & \textcolor{black}{\textbf{22.3}} & \textcolor{black}{25.8} & \textcolor{black}{\textbf{11.8}} \\
Prithvi-EO-2.0-600M & \textcolor{black}{31.1} & \textcolor{black}{\textbf{75.9}} & \textcolor{black}{21.0} & \textcolor{black}{20.5} & \textcolor{black}{\textbf{27.5}} & \textcolor{black}{10.4} \\
\bottomrule
\end{tabular}
\end{table}

\subsubsection{\textcolor{black}{Landslide Detection}}

\textcolor{black}{ Table~\ref{tab:landslide_full_dataset_training} shows the results for the full training set of the L4S task, where only the configurations yielding the best performance for each model are shown.} From the results, Prithvi-EO-2.0-300M, trained with Lovasz loss, outperformed both U-Net models in terms of mIoU, F1 score, and precision by a noticeable margin. Prithvi-EO-2.0-600M, trained with weighted cross-entropy loss, exhibited a slightly lower mIoU of 70.4\% and an F1 score of 58.6\%, but achieved a higher recall than Prithvi-EO-2.0-300M. In summary, Prithvi-EO-2.0-300M demonstrated the best overall performance based on mIoU and F1 score, while Prithvi-EO-2.0-600M showed a higher recall but sacrificed precision.

\begin{table}[htbp]
\centering
\caption{\textcolor{black}{Performance metrics on the test set of L4S for models trained with the} full dataset (100\% or 3799 images).}
\small
\begin{tabular}{@{}l@{\hspace{7.5pt}}c@{\hspace{7.5pt}}c@{\hspace{7.5pt}}c@{\hspace{7.5pt}}c@{\hspace{7.5pt}}c@{\hspace{7.5pt}}c@{}}
\toprule
\textbf{Model} & \textbf{\textcolor{black}{loss}} & \textbf{mIoU} & \textbf{F1} & \textbf{Prec.} & \textbf{Recall} & \textbf{\textcolor{black}{m}Acc} \\
\midrule
U-Net & wCE & 70.4 & 59.7 & 56.2 & 63.6 & 98.4 \\
U-Net++ & wCE & 69.0 & 57.0 & 50.2 & 68.4 & 98.0 \\
Prithvi-EO-2.0-\textcolor{black}{3}00M & Lovasz & \textbf{71.3} & \textbf{60.7} & \textbf{57.1} & 64.7 & \textbf{98.4} \\
Prithvi-EO-2.0-600M & wCE & 70.4 & 58.6 & 51.1 & \textbf{68.6} & 98.2 \\
\bottomrule
\end{tabular}
\label{tab:landslide_full_dataset_training}
\end{table}

\begin{table}[ht]
\centering
\caption{\textcolor{black}{Performance metrics on the test set of L4S for models trained with} a small dataset (\textcolor{black}{\textasciitilde1}\% or \textcolor{black}{50} images).}
\small
\begin{tabular}{@{}l@{\hspace{7.5pt}}c@{\hspace{7.5pt}}c@{\hspace{7.5pt}}c@{\hspace{7.5pt}}c@{\hspace{7.5pt}}c@{\hspace{7.5pt}}c@{}}
\toprule
\textbf{Model} & \textbf{\textcolor{black}{loss}} & \textbf{mIoU} & \textbf{F1} & \textbf{Prec.} & \textbf{Recall} & \textbf{\textcolor{black}{m}Acc} \\
\midrule
U-Net & wCE & \textcolor{black}{59.7} & \textcolor{black}{35.0} & \textcolor{black}{52.5} & \textcolor{black}{27.6} & \textcolor{black}{98.1} \\
U-Net++ & wCE & \textcolor{black}{55.5} & \textcolor{black}{22.3} & \textcolor{black}{43.8} & \textcolor{black}{16.9} & \textcolor{black}{98.0} \\
\textcolor{black}{Prithvi-EO-2.0-}300M & Lovasz & \textcolor{black}{64.5} & \textcolor{black}{43.8} & \textcolor{black}{53.0} & \textcolor{black}{37.3} & \textcolor{black}{98.2} \\
\textcolor{black}{Prithvi-EO-2.0-}600M & wCE & \textbf{\textcolor{black}{67.0}} & \textbf{\textcolor{black}{49.7}} & \textbf{\textcolor{black}{57.2}} & \textbf{\textcolor{black}{44.0}} & \textbf{\textcolor{black}{98.3}} \\
\bottomrule
\end{tabular}
\label{tab:landslide_small_subset_training}
\end{table}

\textcolor{black}{In Table~\ref{tab:landslide_small_subset_training}, we present the results} for models trained on \textcolor{black}{a small subset of the training data (only 50} images). As expected, training on such a small subset led to performance degradation across all models, but the degree of decline varied. Both U-Net and \textcolor{black}{U-Net++} experienced \textcolor{black}{significant} performance drop\textcolor{black}{s}. For instance, U-Net's mIoU decreased from 70.4\% to \textcolor{black}{59.7}\%, and its F1 score fell from 59.7\% to \textcolor{black}{35.0\%}. \textcolor{black}{U-Net++'s performance declined even more sharply, yielding a very low recall score (16.9\%). In contrast,} Prithvi-EO-2.0-300M \textcolor{black}{and Prithvi-EO-2.0-600M showed more stable performance. Although both models experienced some performance drops, they were considerably smaller than those observed for U-Net and U-Net++.} Overall, Prithvi-EO-2.0-600M \textcolor{black}{consistently achieved the best results among all models in the small sample set evaluation. This highlights the advantages of large GFMs in learning generalizable patterns from large-scale datasets, as well as the benefits of specialized pretraining on EO data.}

\begin{figure}[hbt!]
\captionsetup[subfigure]{aboveskip=1pt,belowskip=1pt}
\centering
\begin{subfigure}[a]{0.9\columnwidth}
    \includegraphics[width=\columnwidth]{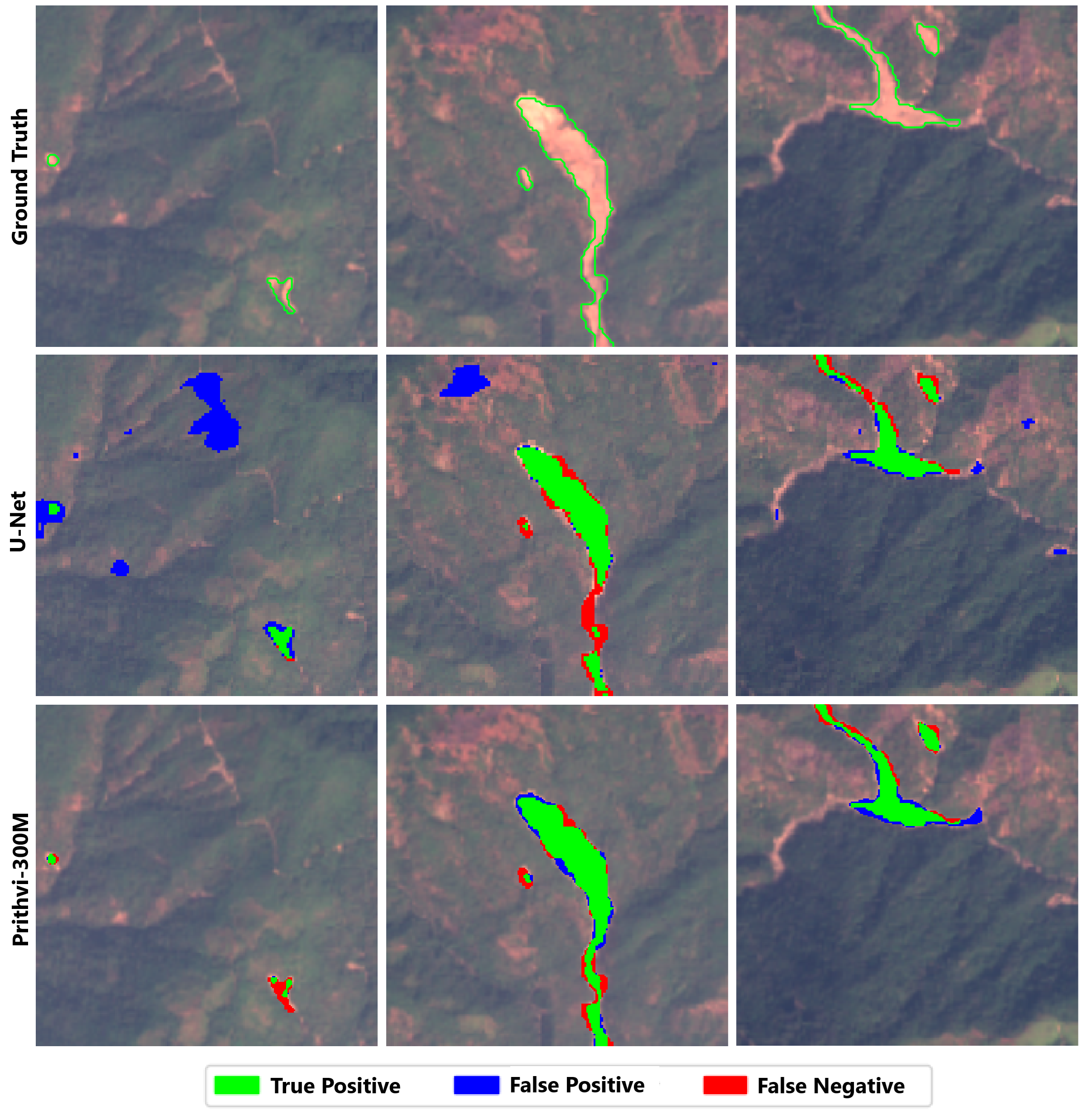}   
    \caption{}
    \label{fig:L4S-full_training} 
\end{subfigure}
\begin{subfigure}[b]{0.9\columnwidth}
    \includegraphics[width=\columnwidth]{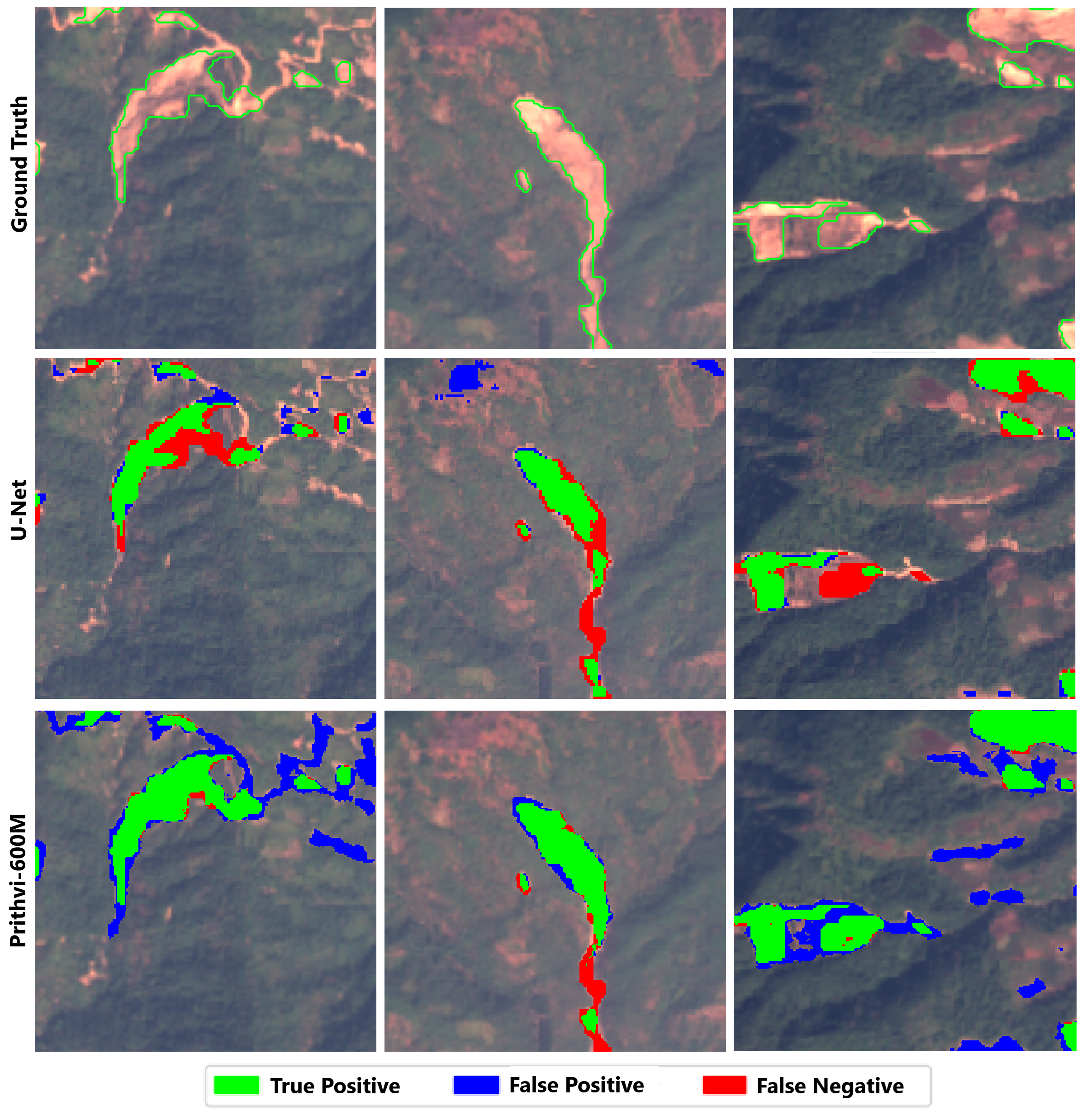}    
    \caption{}
    \label{fig:L4S-small_subset_training}
\end{subfigure}
\caption{\textcolor{black}{Some examples of segmentation results on the test set of L4S to illustrate the models' performance when fine-tuned with (a) the full training set and (b) a small subset (50 images).}} 
\end{figure}

We also selected three pairs of segmentation results from the top-performing Prithvi and U-Net models to conduct a visual comparative analysis of segmentation performance, using sensitivity and specificity statistics. Figure~\ref{fig:L4S-full_training} illustrates the results of U-Net and Prithvi-EO-2.0-300M, both fine-tuned on the full training dataset, while Figure~\ref{fig:L4S-small_subset_training} shows the results for U-Net and Prithvi-EO-2.0-600M, fine-tuned on the smaller subset \textcolor{black}{with 50} images.

\textcolor{black}{From} Figure~\ref{fig:L4S-full_training}, \textcolor{black}{we notice that} Prithvi-EO-2.0-300M exhibits a more conservative detection approach, with fewer False Positives (blue) but a broader coverage of True Positives (green), capturing the target regions with higher accuracy. However, it shows a slightly higher incidence of False Negatives (red) in some cases. In contrast, U-Net results in more False Positives \textcolor{black}{(blue)}, leading to a lower accuracy and an overall \textcolor{black}{lower F1} score. This indicates that Prithvi-EO-2.0-300M prioritizes detection accuracy by effectively reducing False Positives. While this approach may potentially sacrifice recall in certain areas, it achieves the best predictive performance among all models \textcolor{black}{when the full L4S dataset is used}. 


Figure~\ref{fig:L4S-small_subset_training} reveals a different pattern in the segmentation results of U-Net and Prithvi-EO-2.0-600M. U-Net tends to produce more False Negatives, often missing parts of the target regions, \textcolor{black}{resulting in a decrease in recall}. Prithvi-EO-2.0-600M, \textcolor{black}{in comparison}, shows \textcolor{black}{stronger} performance with \textcolor{black}{the detection of substantial} True Positives across the target regions, although it \textcolor{black}{it tends to generate} more False Positives, \textcolor{black}{lowering precision} compared to \textcolor{black}{when the full training dataset is used}. \textcolor{black}{Overall,} Prithvi-EO-2.0-600M \textcolor{black}{remains strong when trained with a small sample size, benefiting from} its enhanced model capacity to generalize \textcolor{black}{to unseen locations} effectively, even with limited training data, prioritizing sensitivity and reducing under-segmentation compared to U-Net.

\subsection{Land cover and crop mapping}

\subsubsection{\textcolor{black}{Multi-Temporal Crop Segmentation in the United States}}

The results \textcolor{black}{in Table ~\ref{tab:multitemporalcropus}} for the crop segmentation task in the United States show that \textcolor{black}{all Prithvi-EO} models outperform the U-Net model in terms of mIoU. \textcolor{black}{Furthermore, the Prithvi-EO-2.0 versions achieved considerably higher mIoU and mAcc values compared to their predecessor.} Among all the Prithvi models, the 600M variant achieved the best performance, with a mIoU of 50.7\% and mAcc of 68.8\%. \textcolor{black}{These results indicate that Prithvi-EO-2.0 was able to capture changes over time in the three-frame inputs required to correctly identify the crops.}

\begin{table}[htbp]
\centering
\caption{Results of the US multi-temporal crop segmentation dataset.}
\label{tab:multitemporalcropus}
\small
\begin{tabular}{@{}lcc@{}}
\toprule
\textbf{Model} & \textbf{mIoU} & \textbf{mAcc} \\
\midrule
U-Net                & 42.6 & 61.9  \\
Prithvi-EO-1.0-100M  & 42.7 & 60.7  \\
Prithvi-EO-2.0-300M  & 48.6 & 66.8  \\
Prithvi-EO-2.0-600M  & \textbf{50.7} & \textbf{68.8}  \\
\bottomrule
\end{tabular}
\end{table}

\subsubsection{\textcolor{black}{Multi-Temporal Land Cover and Crop Classification in Europe}}

Table~\ref{tab:land_cover_fractions_perf} presents the results for \textcolor{black}{the} land cover classification task in Europe. \textcolor{black}{In terms of weighted-averaged F1 score, all Prithvi-EO models outperform the baseline ViViT, with significant margins considering the smaller data subsets. Compared to the other models}, Prithvi-EO-2.0-600M \textcolor{black}{consistently} demonstrates superior performance across all data \textcolor{black}{availability scenarios. Additionally, our new versions of the model surpass Prithvi-EO-1.0 performance.}

Table~\ref{tab:crops_fractions_perf} depicts results on the crop type classification task, where \textcolor{black}{the Prithvi models also outperform the baseline. In this case,} Prithvi-EO-2.0-600M achieves \textcolor{black}{F1 scores} comparable to Prithvi-EO-2.0-300M, showing marginally better results on two data \textcolor{black}{ratios} and slightly lower performance on \textcolor{black}{the other three. B}oth models demonstrate superior performance compared to Prithvi-EO-1.0-100M.

\textcolor{black}{It is worth noting that even though the Prithvi-EO models were pretrained with samples containing four frames, they show robust performance on the Sen4Map tasks, which comprise 12-frame input samples. This indicates they can effectively handle larger time-series inputs.}

\begin{table}[hbtp]
\centering
\caption{\textcolor{black}{Weighted} F1 score and \textcolor{black}{standard deviation} for the land cover classification \textcolor{black}{task}, \textcolor{black}{for different ratios} of the original \textcolor{black}{Sen4Map} dataset.}
\label{tab:land_cover_fractions_perf}
\small
\begin{tabular}{@{}lcc@{\hspace{10pt}}cc@{}}
\toprule
\textbf{\textcolor{black}{Ratio}} & \textbf{\makecell{ViViT\\(Sen4Map)}} &  \textbf{\makecell{Prithvi-EO-\\1.0-100M}} &  \textbf{\makecell{Prithvi-EO-\\2.0-300M}} &  \textbf{\makecell{Prithvi-EO-\\2.0-600M}} \\
\midrule
1.0    & 72.7 (0.3) &      74.5 (0.3) &  76.0 (0.3) &  \textbf{76.1} (0.2) \\
0.5    & 69.9 (0.4) &      72.4 (0.4) &  73.9 (0.4) &  \textbf{74.1} (0.3) \\
0.25   & 65.7 (0.7) &      69.1 (0.5) &  70.7 (0.6) &  \textbf{71.2} (0.5) \\
0.125  & 61.6 (0.5) &      65.3 (0.5) &  67.1 (1.1) &  \textbf{67.8} (1.4) \\
0.0625 & 55.0 (1.3) &      59.4 (1.3) &  61.2 (1.4) &  \textbf{62.3} (0.9) \\
\bottomrule
\end{tabular}
\end{table}

\begin{table}[hbtp]
\centering
\caption{\textcolor{black}{Weighted} F1 score and \textcolor{black}{standard deviation} for the crop type classification task, \textcolor{black}{for different ratios} of the original \textcolor{black}{Sen4Map} dataset.}
\label{tab:crops_fractions_perf}
\small
\begin{tabular}{@{}lcc@{\hspace{10pt}}cc@{}}
\toprule
\textbf{\textcolor{black}{Ratio}} & \textbf{\makecell{ViViT\\(Sen4Map)}} &  \textbf{\makecell{Prithvi-EO-\\1.0-100M}} &  \textbf{\makecell{Prithvi-EO-\\2.0-300M}} &  \textbf{\makecell{Prithvi-EO-\\2.0-600M}} \\
\midrule
1.0    &    81.5 (0.3) &      83.0 (0.3) &  84.4 (0.2) &  \textbf{84.6} (0.2) \\
0.5    &    79.0 (0.3) &      81.5 (0.2) &  \textbf{82.8} (0.2) &  82.6 (0.2)  \\
0.25   &    75.5 (0.4) &      78.6 (0.4) &  80.4 (0.4) &  \textbf{80.5} (0.3)\\
0.125  &    69.7 (1.0) &      74.6 (0.7) &  \textbf{77.3} (0.4) &  76.9 (0.5) \\
0.0625 &    64.5 (1.4) &      70.5 (1.0) &  \textbf{72.2} (1.4) &  70.8 (0.9) \\
\bottomrule
\end{tabular}
\end{table}

\subsubsection{\textcolor{black}{Multi-Temporal Crop Segmentation with PASTIS}}

\textcolor{black}{Table~\ref{tab:pastis_results} shows the results for our experiments with the PASTIS dataset, considering 100\% and 10\% of the training data. Prithvi-EO-2.0-600M achieved the highest mIoU in both scenarios, followed by the baseline model. As expected, all models experienced a significant drop in performance when trained on only 10\% of the data, which corresponds to approximately 200 images.}

\textcolor{black}{DOFA and Satlas present lower scores compared to the baseline, which indicates that the lack of temporal modeling in the architectures and pretraining datasets may limit their ability to handle long sequences. In contrast, Presto showed the worst results. We attribute this to the input image size, which may be large for a model originally designed for single-pixel time-series data.}

\textcolor{black}{These results suggest that Prithvi-EO-2.0 was able to learn with long sequences, better capturing the temporal patterns compared to the other GFMs, highlighting the value of its pretraining data and architecture.}

\begin{table}[bt]
\centering
\caption{\textcolor{black}{Results on crop segmentation with the PASTIS dataset. We show the mIoU for the models trained with 100\% and 10\% of the data.}}
\label{tab:pastis_results}
\small
\begin{tabular}{@{}>{\color{black}}l >{\color{black}}c >{\color{black}}c@{}}
\toprule
\textbf{Model} & \textbf{10\%} & \textbf{100\%} \\
\midrule
DOFA                 & 20.4 & 37.0 \\
Satlas               & 15.8 & 31.5 \\
Presto               & 20.1 & 28.1 \\
U‑TAE (baseline)     & 36.9 & 52.9 \\
Prithvi-EO-2.0‑300M  & 36.3 & 51.6 \\
Prithvi-EO-2.0‑600M  & \textbf{37.4} & \textbf{53.4} \\
\bottomrule
\end{tabular}
\end{table}

\subsection{Ecosystem dynamics}

\subsubsection{\textcolor{black}{Above Ground Biomass Estimation}}

\textcolor{black}{Table~\ref{tab:biomassters_performance} shows the test set results for Prithvi-EO-2.0-300M and the baseline U-Net. The best performance for Prithvi-EO is achieved using 12 timestamps of Sentinel-2 MSI as inputs and LoRA. Performance for the same model fine-tuned on 5\% of the training set (348 images) only differs by 10.36\% from the best result on the full training set (6,951 images), as seen in Table~\ref{tab:biomassters_subset_performance}.}

\begin{figure}[t] 
    \begin{subfigure}[a]{\columnwidth}
         \centering
         \includegraphics[
            width=0.8\textwidth,
            trim=1cm 0.5cm 3cm 1.7cm,
            clip
        ]{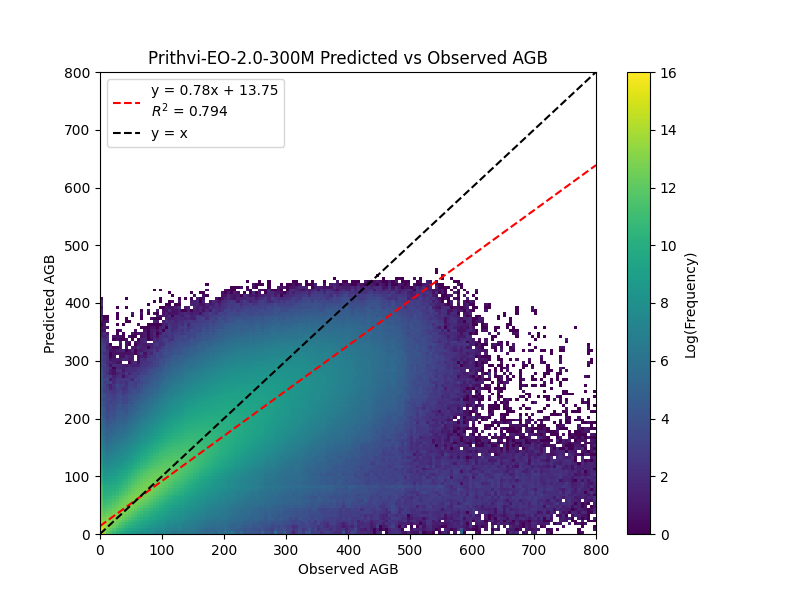}
         \caption{}
     \end{subfigure}
     \hfill
     \begin{subfigure}[b]{\columnwidth}
         \centering
         \includegraphics[
            width=0.8\textwidth,
            trim=1cm 0.5cm 3cm 1cm,
            clip
        ]{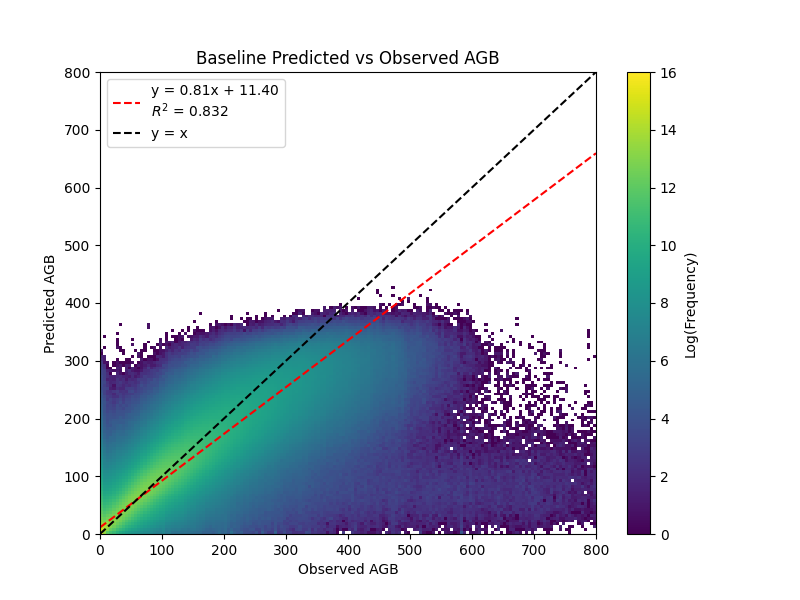}
         \caption{}
     \end{subfigure}
\caption{\textcolor{black}{Comparison of AGB predictions versus observed AGB on all inputs in the test set for (a) fine-tuned Prithvi-EO-2.0-300M and (b) baseline model. Prithvi-EO-2.0-300M is fine-tuned and tested using 12 timestamps of Sentinel-2 data. A log scale is used for visualization.}} 
\label{fig:biomassters_pred_scatter} 
\end{figure}

Figure~\ref{fig:biomassters_pred_scatter} shows the $R^2$ scores of the best performing Prithvi\textcolor{black}{-EO-300M} and the baseline \textcolor{black}{model across} all AGB targets in the test set. The baseline model \textcolor{black}{exhibits} a tighter distribution around the identity line \textcolor{black}{and} more accurate prediction\textcolor{black}{s} of high AGB values \textcolor{black}{at the range} 300~-~400. However, both models underestimate higher AGB values ($>$~400).

\begin{table}[htbp]
\centering
\caption{\textcolor{black}{RMSE values for Prithvi-EO-2.0-300M on BioMassters varying input configuration (T = input length).}}
\label{tab:biomassters_performance}
\small
\begin{tabular}{@{}lcc@{}}
\toprule
\textbf{\textcolor{black}{Data}} & \textbf{\textcolor{black}{T}} & \textbf{\textcolor{black}{Test Set RMSE}} \\
\midrule
\textcolor{black}{S2 (LoRA)}& \textcolor{black}{4} & \textcolor{black}{35.47} \\
\textcolor{black}{S2 + S1} & \textcolor{black}{4} & \textcolor{black}{38.67} \\
\textcolor{black}{S2 (LoRA)}& \textcolor{black}{12} & \textcolor{black}{33.40} \\
\textcolor{black}{S2 + S1} & \textcolor{black}{12} & \textcolor{black}{36.48} \\
\textcolor{black}{All S2 Bands + S1} & \textcolor{black}{12} & \textcolor{black}{36.25} \\
\midrule
\textcolor{black}{baseline} & & \textbf{\textcolor{black}{27.49}} \\
\bottomrule
\end{tabular}
\end{table}

\begin{table}[htbp]
\centering
\caption{\textcolor{black}{RMSE values of Prithvi-EO-2.0-300M model on BioMassters test set across different data subsets using 12 monthly observations of Sentinel-2 and LoRA.}}
\label{tab:biomassters_subset_performance}
\small
\begin{tabular}{@{}lc@{}}
\toprule
\textbf{\textcolor{black}{Subset}} & \textbf{\textcolor{black}{Test Set RMSE}} \\
\midrule
\textcolor{black}{50\%} & \textcolor{black}{34.31} \\
\textcolor{black}{20\%} & \textcolor{black}{34.89} \\
\textcolor{black}{10\%} & \textcolor{black}{36.77} \\
\textcolor{black}{5\%} & \textcolor{black}{38.06} \\
\bottomrule
\end{tabular}
\end{table}

\begin{figure*}[t] 
\centering 
\includegraphics[width=1.35\columnwidth]{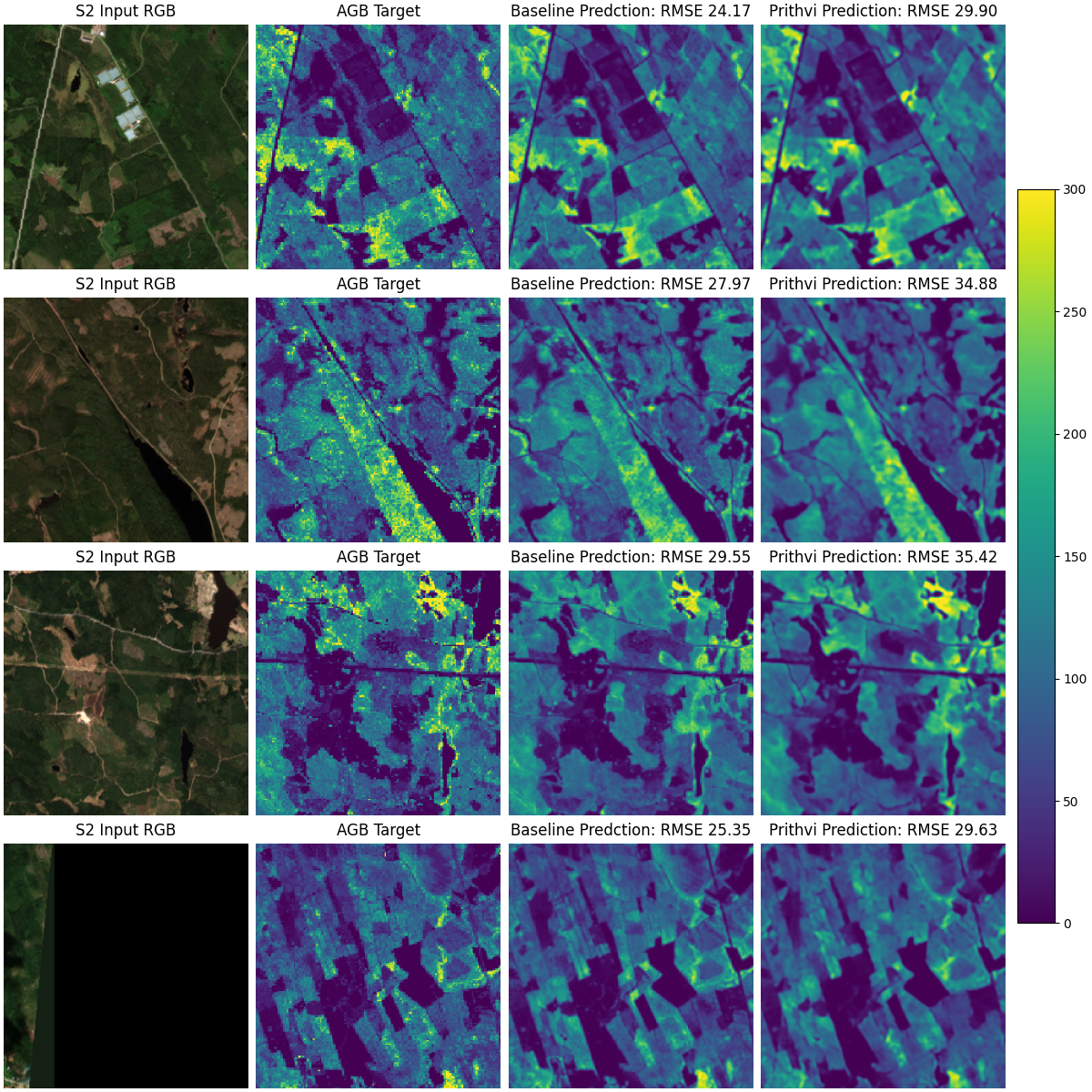} 
\caption{Comparison of fine-tuned Prithvi and baseline model prediction for \textcolor{black}{4} images and their corresponding AGB target pairs from the test set. Columns from left to right: RGB visualization of Sentinel-2 scene from \textcolor{black}{the same month as the AGB observation}, target AGB, baseline model prediction, and fine-tuned Prithvi prediction. As high-quality Sentinel-2 data does not exist for all AGB targets, some S2 inputs, such as the one in row \textcolor{black}{4}, are either obscured by clouds or contain partial data.} 
\label{fig:biomassters_pred_compare} 
\end{figure*}

\textcolor{black}{A} qualitative \textcolor{black}{comparison} of selected outputs \textcolor{black}{depicted} in Figure~\ref{fig:biomassters_pred_compare} \textcolor{black}{reveals} that \textcolor{black}{the} baseline model and fine-tuned Prithvi \textcolor{black}{generate visually similar outputs} for some input scenes. Both models \textcolor{black}{produce smoother} predictions than the observed AGB values. \textcolor{black}{The baseline model appears to predict higher AGB values more accurately and with more granularity than the fine-tuned Prithvi-EO model}.

\textcolor{black}{This result is consistent with existing findings that a combination of SAR and multispectral imagery is superior to multispectral imagery alone for biomass estimation, especially in locations with high biomass  \cite{shaoEstimatingForestAboveground2016} \cite{nascetti2023biomassters}. Optical data is prone to saturation in high biomass regions where reflectance values no longer increase proportionally with biomass. This is because multispectral sensors are only capable of sensing the reflectance of the top layer of canopy, and so cannot be used to accurately estimate biomass of vegetation obscured by a full canopy. In these regions, SAR sensors are capable of providing information about the physical structure of vegetation because their microwave signals can penetrate the canopy and interact with branches, stems, and trunks. This penetration capability gives SAR the ability to sense the vertical and structural variability of biomass, which is not directly captured by optical multispectral sensors.}

\textcolor{black}{Our results show that Prithvi approaches the performance of multimodal models on this task using solely multispectral imagery, and is capable of adapting to and benefiting from longer time series of inputs than used in pretraining. This is achieved with an efficient fine-tuning technique, lowering training time and cost. Incorporating SAR inputs as additional bands to the Prithvi-EO encoder did not result in increased performance for biomass estimation in our experiment setting. We note that adapting to the new SAR inputs can be challenging for Prithvi-EO given the fundamental differences between optical and SAR data. Therefore, other methods of incorporating different modalities, such as fusing outputs of a separate SAR encoder with the outputs of the pretrained Prithvi-EO encoder, could improve performance for this task.}

\subsubsection{\textcolor{black}{Estimation of Gross Primary Productivity (GPP) at Globally Distributed Sites}}

\begin{figure}[htbp]
    \centering
    \subfloat[\textcolor{black}{Train (-2018)}]{%
        \includegraphics[width=0.485\linewidth]{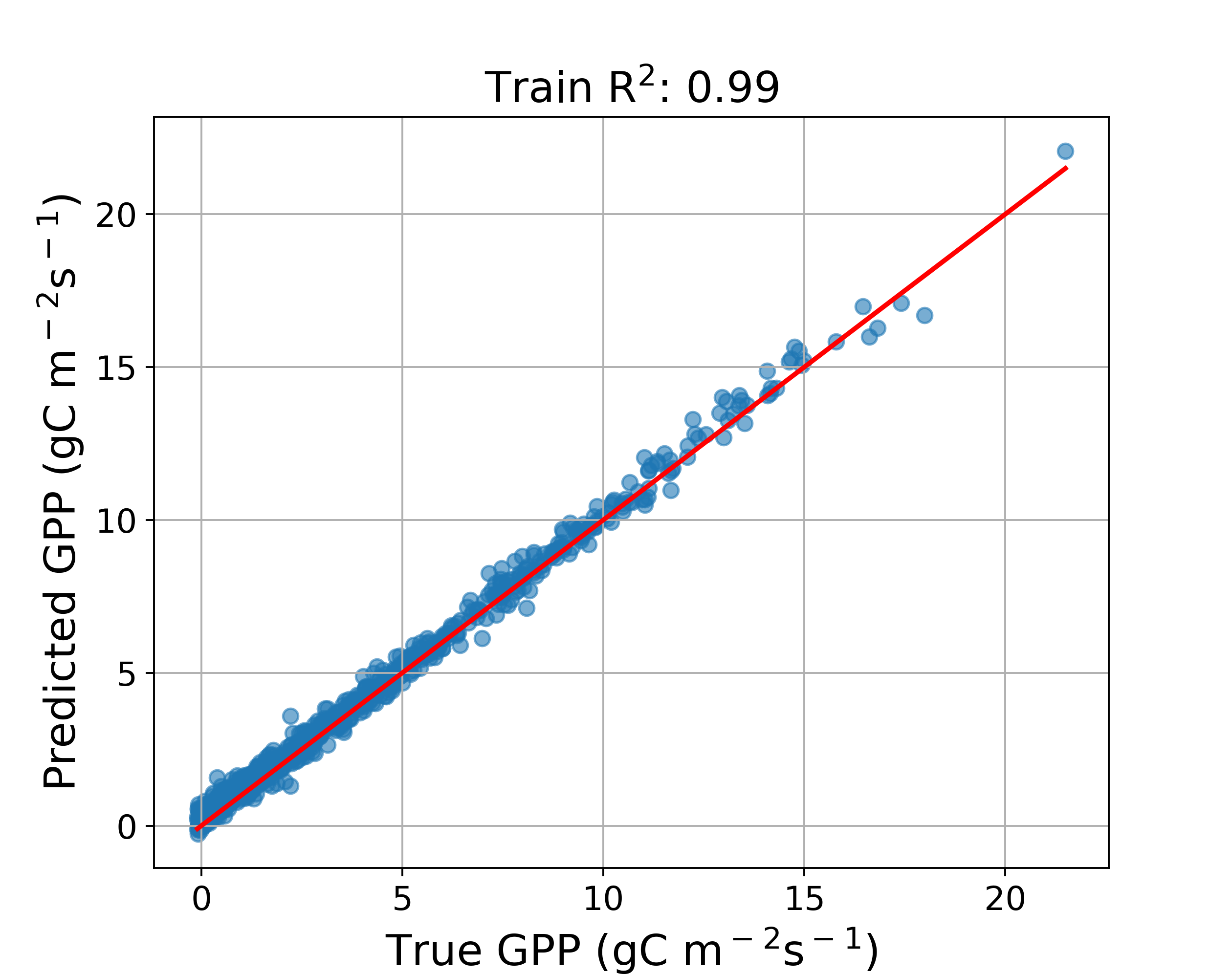}%
        \label{fig:flux_train2018}
    }
    \hfill
    \subfloat[\textcolor{black}{Test (2018)}]{%
        \includegraphics[width=0.485\linewidth]{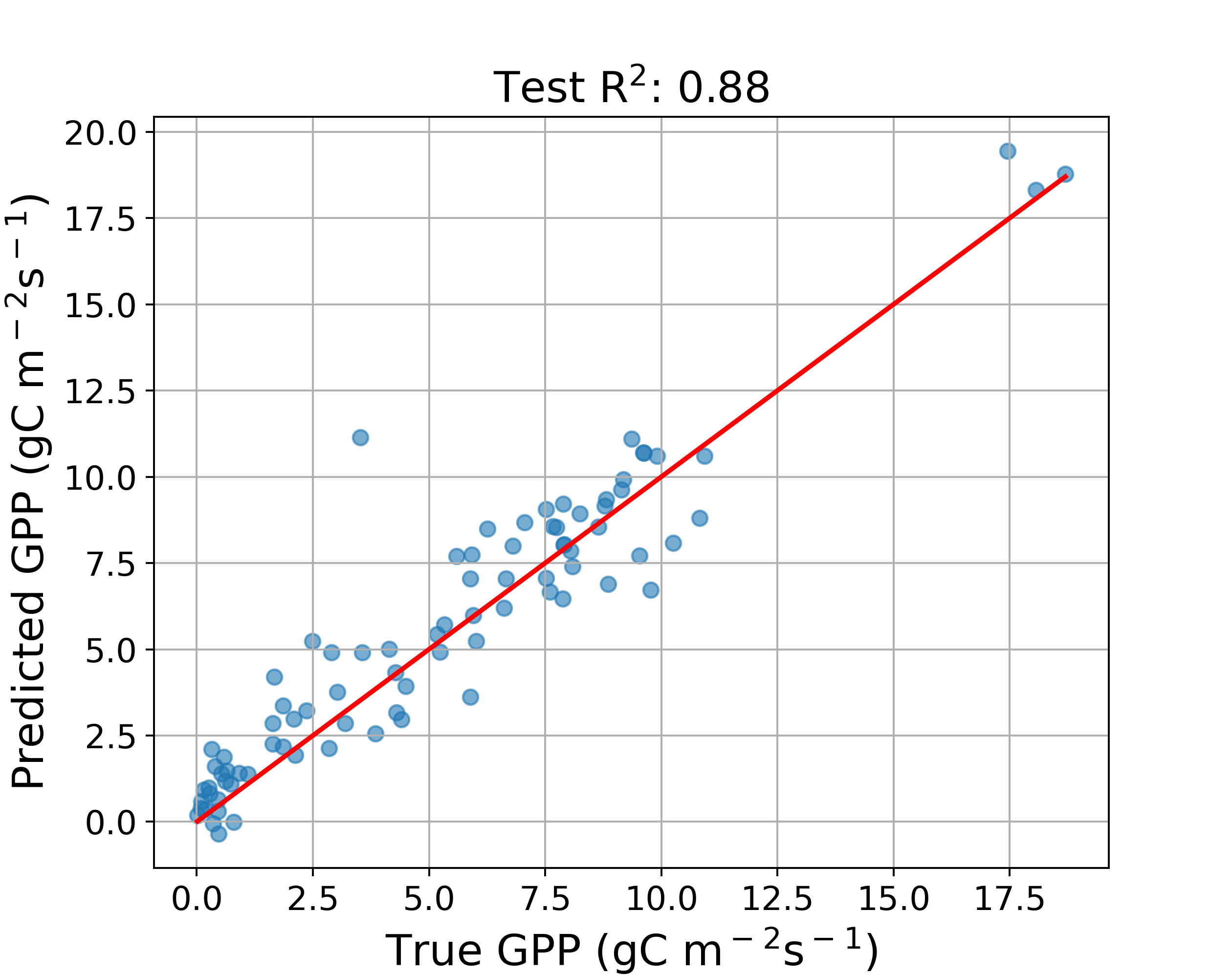}%
        \label{fig:flux_test2018}
    }

    \subfloat[\textcolor{black}{Train (-2019)}]{%
        \includegraphics[width=0.485\linewidth]{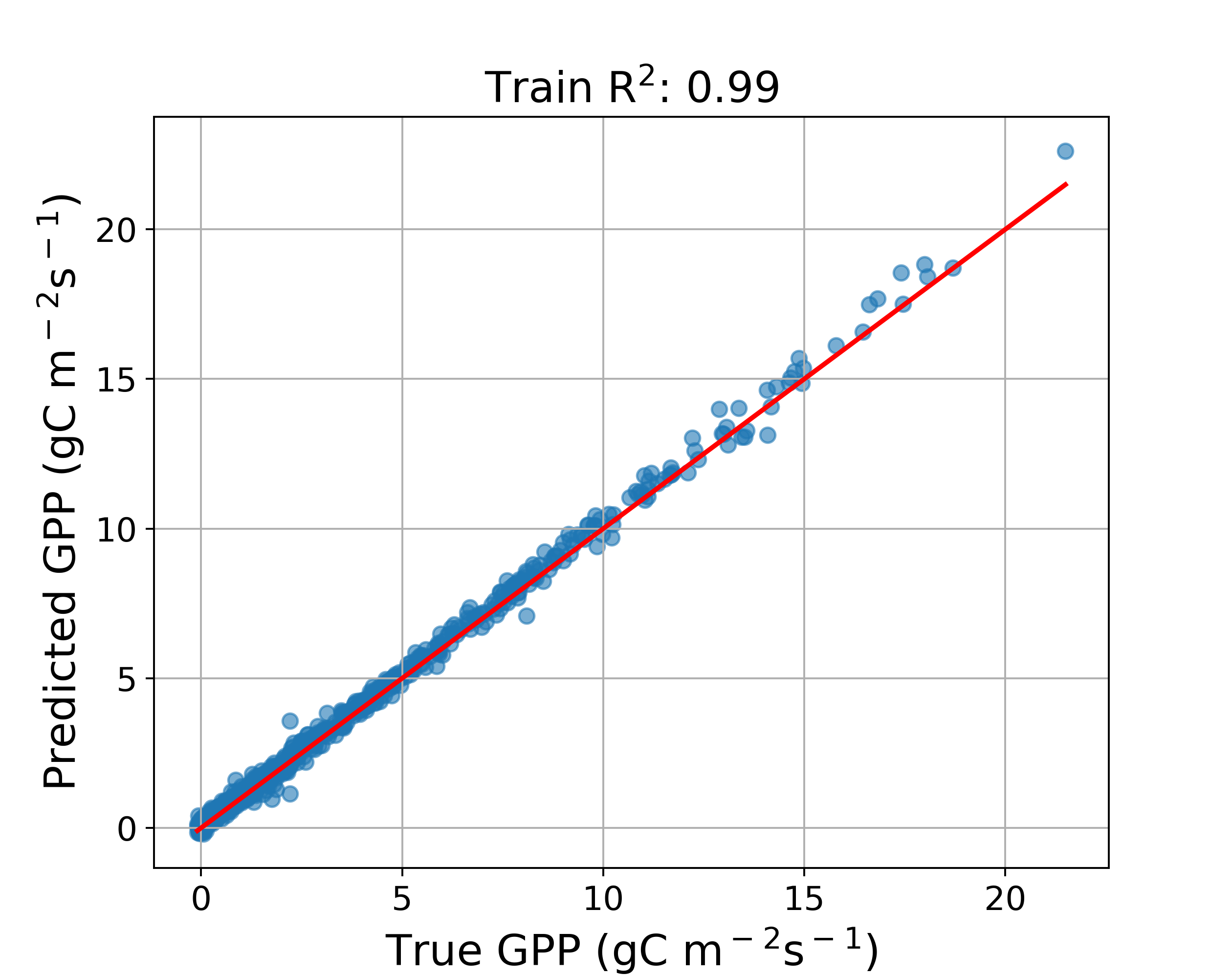}%
        \label{fig:flux_train2019}
    }
    \hfill
    \subfloat[\textcolor{black}{Test (2019)}]{%
        \includegraphics[width=0.485\linewidth]{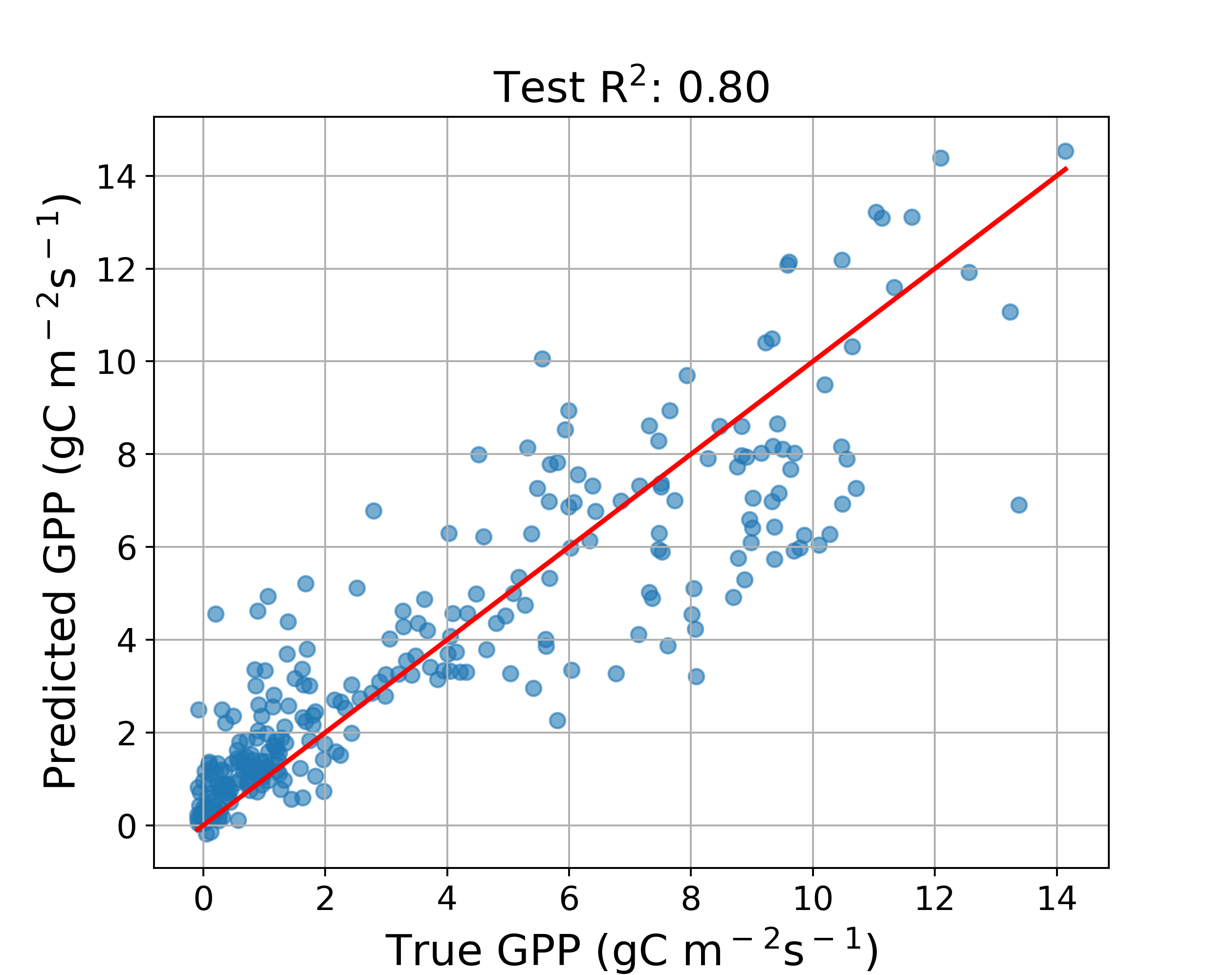}%
        \label{fig:flux_test2019}
    }

    \subfloat[\textcolor{black}{Train (-2020)}]{%
        \includegraphics[width=0.485\linewidth]{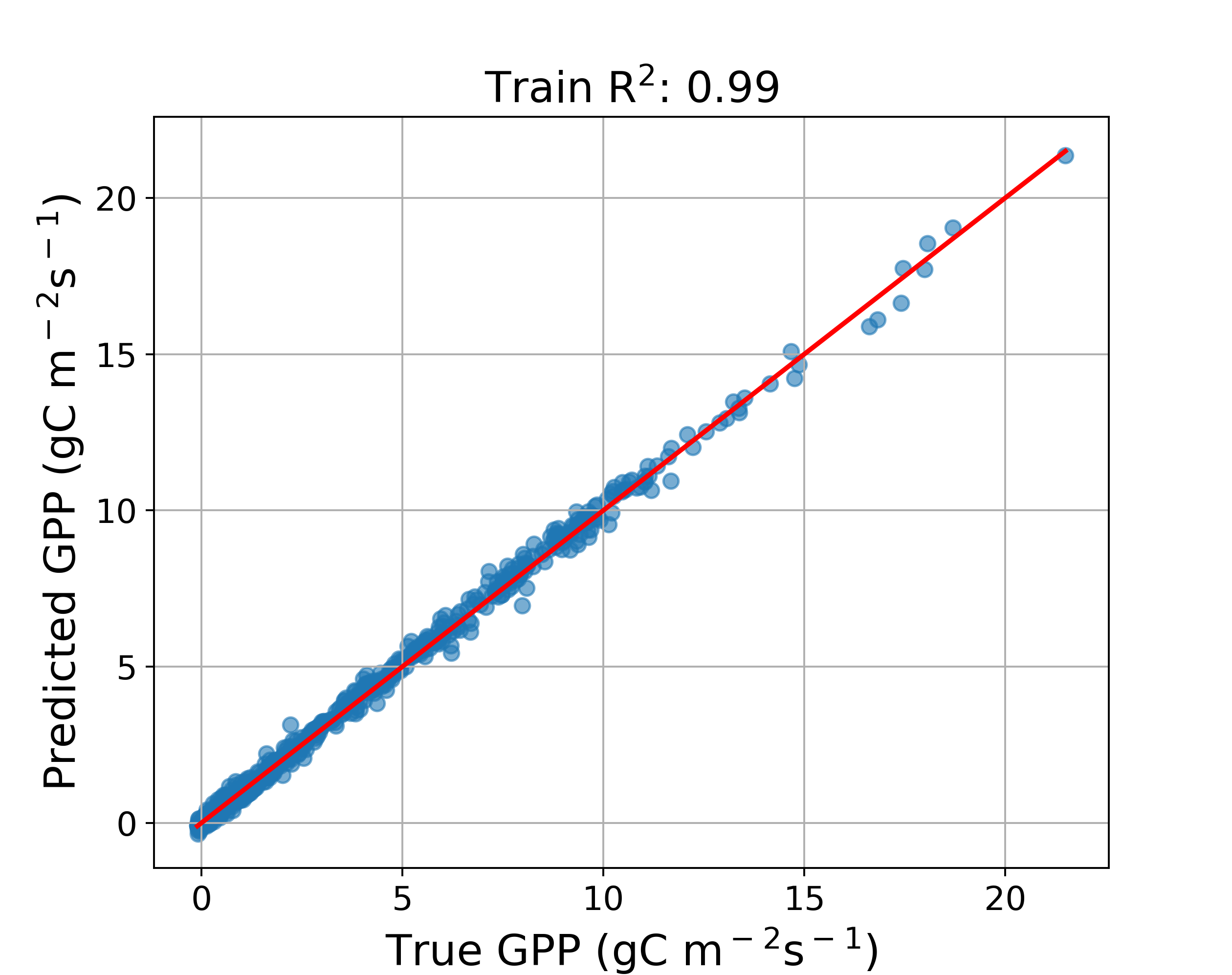}%
        \label{fig:flux_train2020}
    }
    \hfill
    \subfloat[\textcolor{black}{Test (2020)}]{%
        \includegraphics[width=0.485\linewidth]{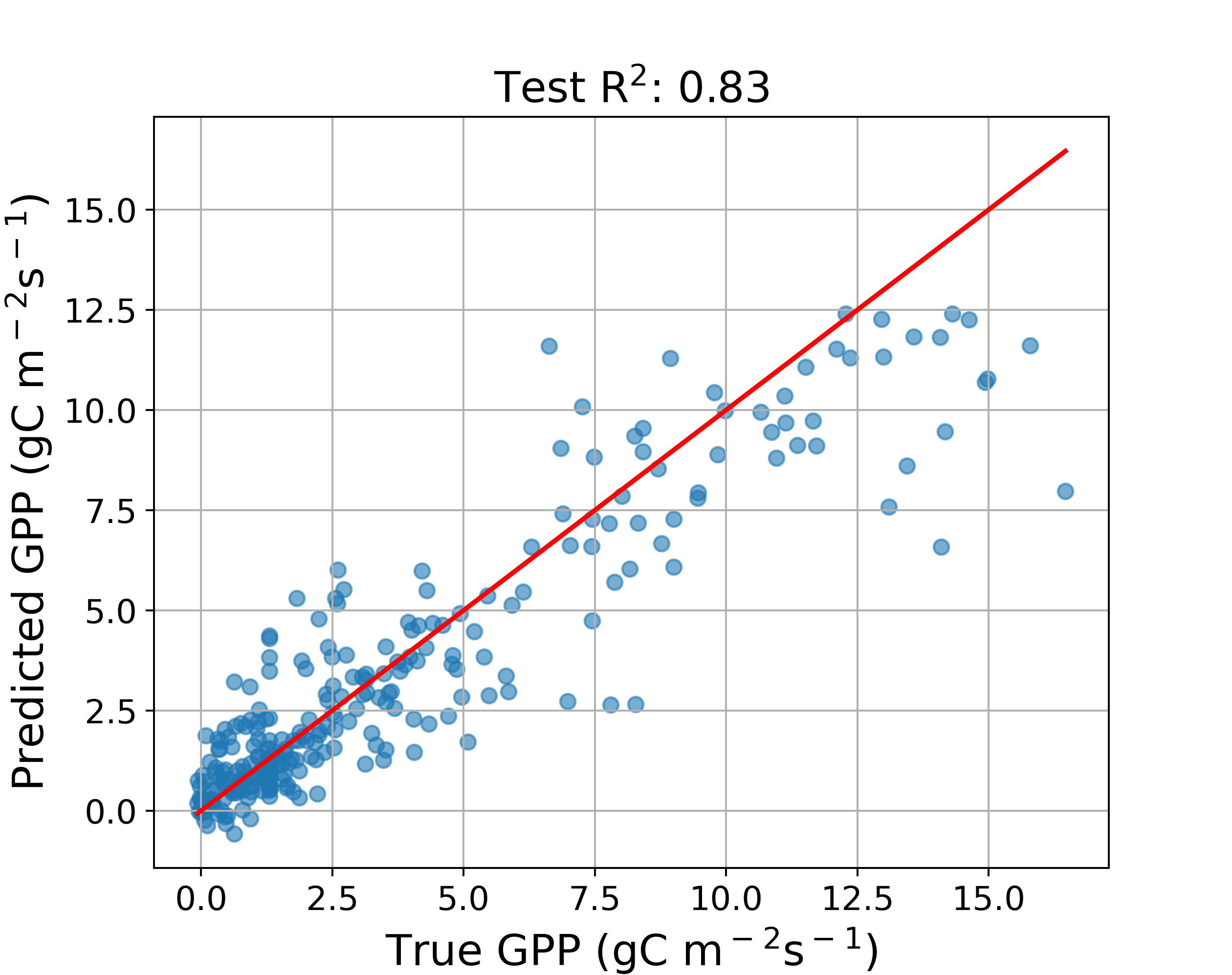}%
        \label{fig:flux_test2020}
    }

    \subfloat[\textcolor{black}{Train (-2021)}]{%
        \includegraphics[width=0.485\linewidth]{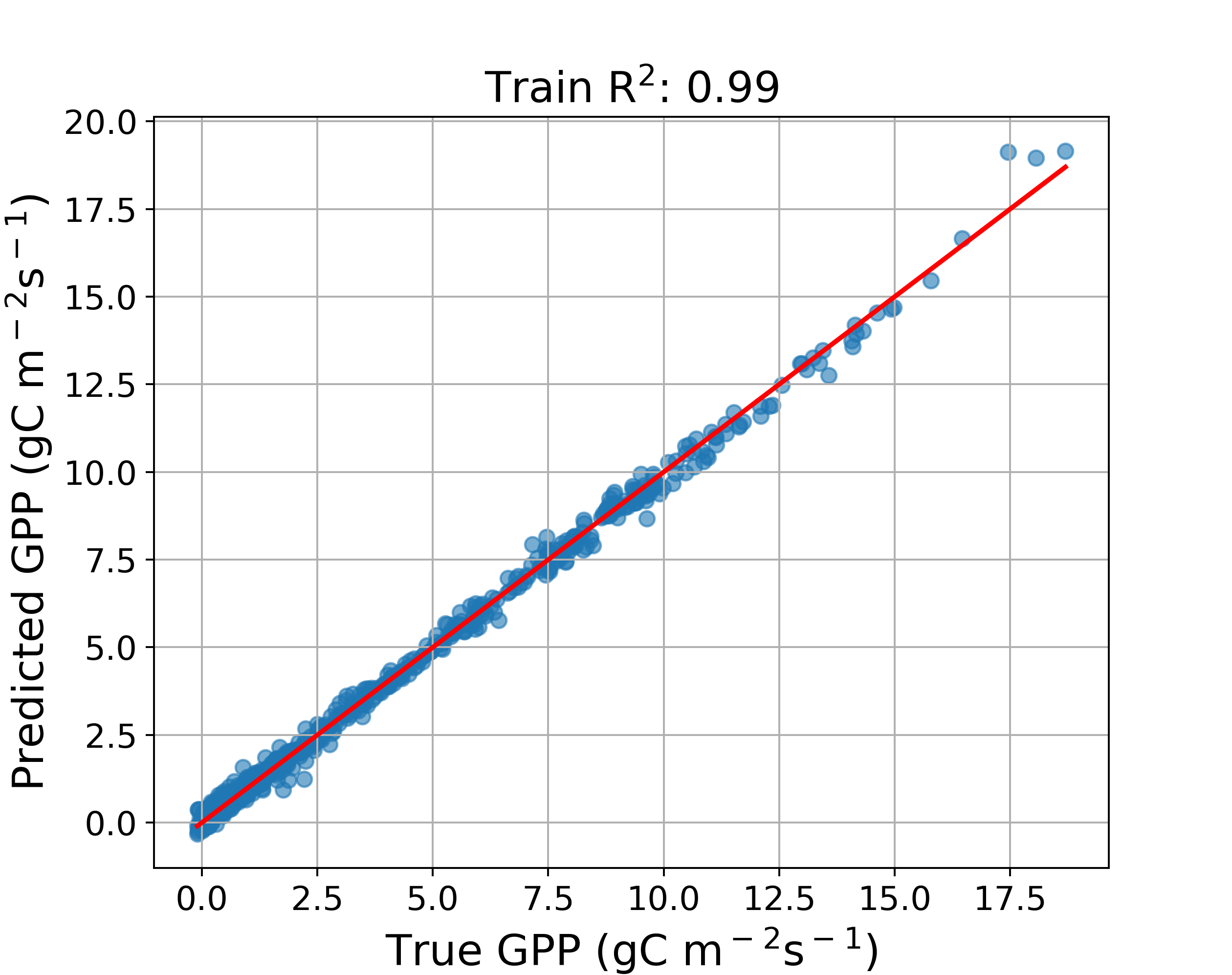}%
        \label{fig:flux_train2021}
    }
    \hfill
    \subfloat[\textcolor{black}{Test (2021)}]{%
        \includegraphics[width=0.485\linewidth]{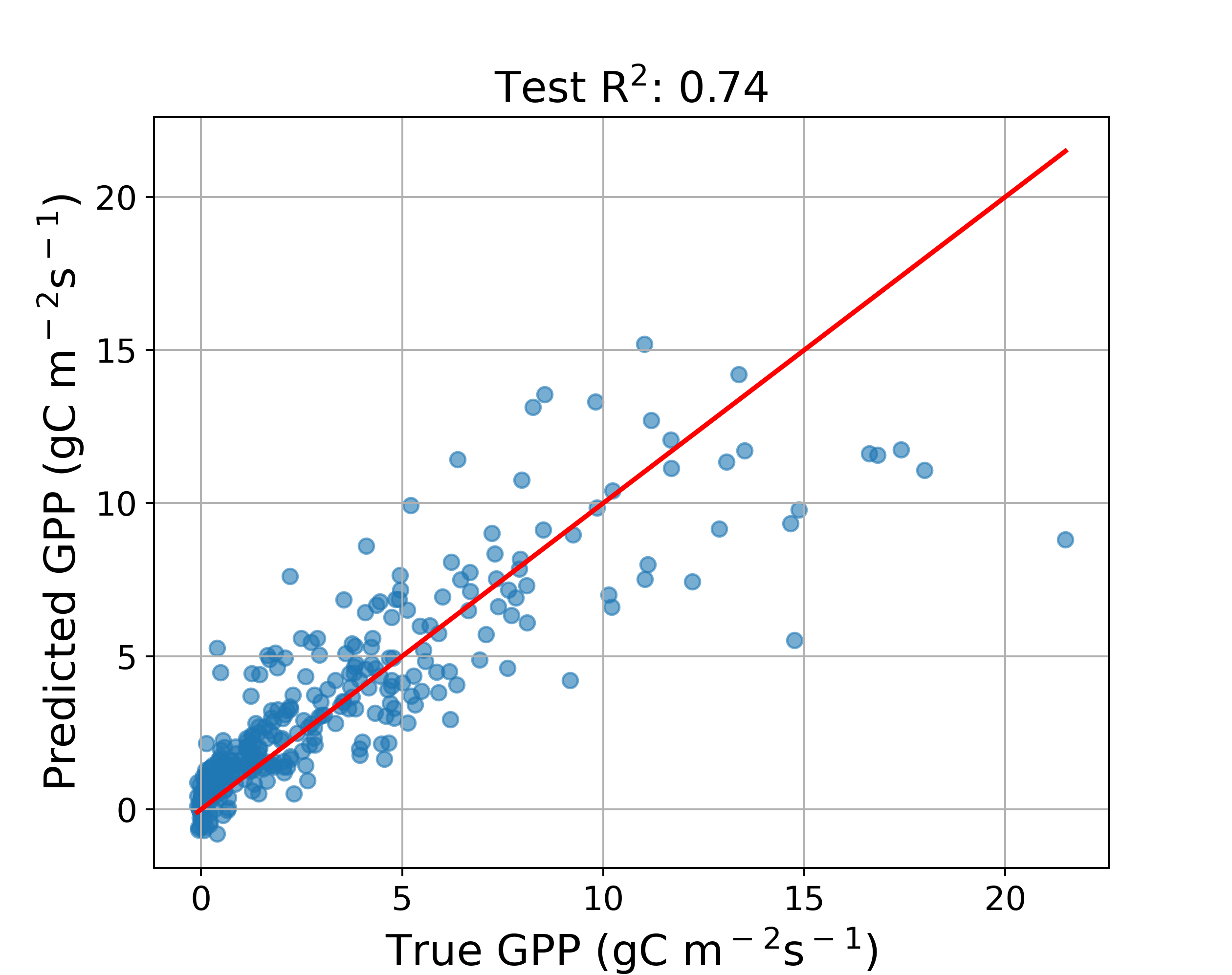}%
        \label{fig:flux_test2021}
    }

    \caption{Train-test set $R^2$ for GPP flux estimation using leave-one-year-out cross-validation over 975 instances of HLS and MERRA-2 data using Prithvi-EO-2.0-600M-TL. \textcolor{black}{Each row shows one of the leave-one-year-out train-test split.}}
    \label{fig:flux_r2_all}
\end{figure}

\textcolor{black}{We provide the $R^2$ values on the GPP estimation task for all tested models} in Table~\ref{tab:R2-GPP-columns}. \textcolor{black}{The r}esults show that Prithvi-EO-2.0 models performed better than the random forest and XGBoost models without vegetation \textcolor{black}{indices} with up-to 20\% improvement on $R^2$. The Prithvi-EO-2.0-600M\textcolor{black}{-TL} version achieved the highest \textcolor{black}{scores} among all models across all testing years. Notably, while including vegetation \textcolor{black}{indices} marginally increased the performance of random forest and XGBoost models, Prithvi-EO-2.0 does not use VIs as inputs. This indicates the superior ability of Prithvi-EO-2.0 to efficiently utilize spatio-spectral information.

\begin{table}[htbp]
\centering
\caption{$R^2$ analysis of baseline model Random Forest vs Prithvi (VIs - vegetation indices, Prithvi Models do not use additional VI) using a leave-one year out cross validation approach over \textcolor{black}{37} globally distributed flux towers (975 samples). The years shown here indicate the test set. \textcolor{black}{TL: pretrained with temporal and location embeddings.}}
\label{tab:R2-GPP-columns}
\small
\begin{tabular}{@{}lcccc@{}}
\toprule
\textbf{Model} & \textbf{2018} & \textbf{2019} & \textbf{2020} & \textbf{2021} \\
\midrule
RF (-VIs) & 0.68 & 0.78 & 0.70 & 0.64 \\
RF (+VIs) & 0.72 & 0.77 & 0.73 & 0.66 \\
XGBoost (-VIs) & 0.66 & 0.75 & 0.69 & 0.60 \\
XGBoost (+VIs) & 0.77 & 0.77 & 0.77 & 0.60 \\
\textcolor{black}{ResNet18} & \textcolor{black}{0.76} & \textcolor{black}{0.81} & \textcolor{black}{0.73} & \textcolor{black}{0.70} \\
Prithvi-EO-2.0-300M & 0.84 & 0.77 & 0.77 & 0.70 \\
Prithvi-EO-2.0-300M-TL & 0.84 & 0.76 & 0.74 & 0.69 \\
Prithvi-EO-2.0-600M & 0.75 & 0.77 & 0.80 & 0.70 \\
Prithvi-EO-2.0-600M-TL & \textbf{0.88} & \textbf{0.80} & \textbf{0.83} & \textbf{0.74} \\
\bottomrule
\end{tabular}
\end{table}

Both random forest and XGBoost baselines use spatial averages of HLS image tiles (1.5~$\times$~1.5~$km^2$) around the tower location, which reflect the current state-of-the-art eddy covariance upscaling approaches using coarse-resolution (e.g.\textcolor{black}{,} 500~m to 5~km) satellite data \cite{kang2025cedar, X-Base}. This method overlooks the spatial heterogeneity of the areas surrounding flux towers and the dynamic nature of the flux measurement footprint. As many sites are situated in heterogeneous landscapes, featuring multiple land cover types and diverse vegetation conditions, the mismatch between the fixed satellite pixel grid and the measurement footprint introduce substantial biases to the upscaling estimation \cite{Chu2021,jung2020scaling}.

Prithvi-EO benefits from the detailed spatial information of the HLS data, thereby enhancing the representation of remote sensing data within the flux measurement footprints. \textcolor{black}{Notably, while the ResNet model also leverages spatial information from HLS images, its performance is consistently lower than Prithvi-EO (i.e.\textcolor{black}{,} average $R^2$ of 0.75 for ResNet vs. 0.81 for Prithvi-EO-2.0-600M-TL). The superior performance of Prithvi-EO demonstrates the advantage of extracting generalizable spatial context by pretraining on massive satellite datasets and incorporating temporal and location embeddings.} Furthermore, the proposed fine-tuning approach using Prithvi-EO-2.0 and MERRA-2 data shows promise for expanding the study to additional flux tower sites and extrapolating GPP beyond the flux network to derive a key climatic variable.

\section{Conclusion}
We presented Prithvi-EO-2.0, the second iteration of the Prithvi-EO family. Prithvi-EO-2.0 showed strong performance in both standard benchmarking experiments at different resolutions, where it improved up-to 8\% its predecessor and outperformed some of the most popular GFMs on GEO-Bench datasets~\cite{GEOBench}. Prithvi-EO-2.0 also showed state-of-the-art results across a range of SME-led real-world applications in disaster response, land cover and crop mapping, and ecosystem dynamics monitoring. \textcolor{black}{As future work, comparisons with multimodal GFMs could help identify the benefits of multimodal pretraining for tasks that include multiple modalities as well as the limitations of incorporating additional data into Prithvi at the fine-tuning stage.} Key to the success of this multi-disciplinary cross-institution project was developing a carefully-crafted unbiased multi-temporal dataset for pretraining, and the transparent and thorough evaluation of many aspects of Prithvi-EO-2.0. The project exemplifies the Trusted Open Science Pledge that all project partners subscribed to. In this spirit, we release our models with a permissive license in two different sizes (300M and 600M) on Hugging Face\footnote{\url{https://huggingface.co/ibm-nasa-geospatial/Prithvi-EO-2.0}}. To maximize community impact and \textcolor{black}{reduce} barriers to entry, we have also onboarded the models into TerraTorch\footnote{\url{https://github.com/NASA-IMPACT/Prithvi-EO-2.0}}, \textcolor{black}{enabling} easy customization and adoption.

\section*{Acknowledgments}
This work is supported by NASA Grant 80MSFC22M004. We want to express our gratitude to Hugging Face for hosting Prithvi, associated demos, and the corresponding datasets for fine-tuning. The authors gratefully acknowledge the Gauss Centre for Supercomputing e.V. (\url{www.gauss-centre.eu}) for funding this project by providing computing time through the John von Neumann Institute for Computing (NIC) on the GCS Supercomputer JUWELS~\cite{JUWELS} at Jülich Supercomputing Centre (JSC). \textcolor{black}{Y. Kang acknowledges additional support from a NASA award 80NSSC24K1562.}

\bibliographystyle{IEEEtran}
\bibliography{IEEEabrv,sn-bibliography_apa}
\clearpage

\begin{appendices}

\section{\textcolor{black}{GEO-Bench detailed results}}

\textcolor{black}{Here we present the detailed GEO-Bench evaluation results for all models. The tables report the mean, standard deviation, maximum and minimum metrics on the test set, computed over ten repeated runs using the best hyperparameters for all models. These results correspond to the boxplots shown in Figure\ref{fig:class_seg_dist_geobench}}.
\vspace{0.815cm}

\renewcommand{\thetable}{A\arabic{table}}
\setcounter{table}{0}
\begin{table}[h!]
\caption{\textcolor{black}{Detailed results for GEO-Bench classification datasets: \textit{m-bigearthnet}, \textit{m-brick-kiln}, and \textit{m-eurosat}.}}
\label{tab:geobench_cls_summary_1}
\begin{tabular}{@{}l@{\hspace{8pt}}rcccc@{}}
\toprule
 & & \multicolumn{4}{c}{Accuracy [\%]} \\
\midrule
dataset & model & mean & std & max & min \\
\midrule
\multirow[t]{13}{45pt}{m-bigearthnet (F1 score)} & DINO-Resnet50 & 67.37 & 0.52 & 68.17 & 66.66 \\
 & DOFA-ViT-300M & 68.58 & 0.22 & 68.77 & 68.14 \\
 & DeCUR-Resnet50 & 67.54 & 0.66 & 68.68 & 66.53 \\
 & MOCO-Resnet50 & 66.56 & 0.57 & 67.11 & 65.08 \\
 & Prithvi-EO-1.0-100M & 66.82 & 0.39 & 67.48 & 66.28 \\
 & Prithvi-EO-2.0-100M & 69.18 & 0.18 & 69.37 & 68.84 \\
 & Prithvi-EO-2.0-100M-TL & 68.96 & 0.37 & 69.39 & 68.09 \\
 & Prithvi-EO-2.0-300M & 70.19 & 0.28 & 70.76 & 69.81 \\
 & Prithvi-EO-2.0-300M-TL & 70.33 & 0.47 & 71.15 & 69.48 \\
 & Prithvi-EO-2.0-600M & 70.73 & 0.42 & 71.17 & 69.72 \\
 & Prithvi-EO-2.0-600M-TL & 70.93 & 0.22 & 71.36 & 70.68 \\
 & Satlas-Swin-100M & 69.63 & 0.50 & 70.40 & 68.65 \\
 & ScaleMAE-ViT-300M & 62.80 & 0.68 & 63.83 & 61.77 \\
\midrule
\multirow[t]{13}{45pt}{m-brick-kiln} & DINO-Resnet50 & 98.43 & 0.30 & 98.90 & 97.90 \\
 & DOFA-ViT-300M & 98.22 & 0.37 & 98.60 & 97.60 \\
 & DeCUR-Resnet50 & 98.27 & 0.30 & 98.90 & 97.80 \\
 & MOCO-Resnet50 & 98.19 & 0.29 & 98.80 & 97.80 \\
 & Prithvi-EO-1.0-100M & 98.24 & 0.34 & 98.60 & 97.60 \\
 & Prithvi-EO-2.0-100M & 98.25 & 0.28 & 98.50 & 97.70 \\
 & Prithvi-EO-2.0-100M-TL & 98.56 & 0.20 & 98.90 & 98.30 \\
 & Prithvi-EO-2.0-300M & 98.31 & 0.17 & 98.50 & 98.00 \\
 & Prithvi-EO-2.0-300M-TL & 98.43 & 0.26 & 98.70 & 97.90 \\
 & Prithvi-EO-2.0-600M & 98.79 & 0.13 & 99.00 & 98.60 \\
 & Prithvi-EO-2.0-600M-TL & 98.61 & 0.19 & 98.90 & 98.20 \\
 & Satlas-Swin-100M & 98.41 & 0.19 & 98.60 & 98.10 \\
 & ScaleMAE-ViT-300M & 98.05 & 0.41 & 98.40 & 97.10 \\
\midrule
\multirow[t]{13}{45pt}{m-eurosat} & DINO-Resnet50 & 96.79 & 0.60 & 97.40 & 95.90 \\
 & DOFA-ViT-300M & 96.18 & 0.79 & 97.00 & 94.90 \\
 & DeCUR-Resnet50 & 97.61 & 0.30 & 98.00 & 97.10 \\
 & MOCO-Resnet50 & 98.16 & 0.36 & 98.80 & 97.60 \\
 & Prithvi-EO-1.0-100M & 92.07 & 0.74 & 93.30 & 90.60 \\
 & Prithvi-EO-2.0-100M & 95.22 & 0.39 & 95.90 & 94.80 \\
 & Prithvi-EO-2.0-100M-TL & 94.72 & 0.39 & 95.40 & 94.10 \\
 & Prithvi-EO-2.0-300M & 94.97 & 0.50 & 95.70 & 94.40 \\
 & Prithvi-EO-2.0-300M-TL & 94.86 & 0.40 & 95.20 & 93.90 \\
 & Prithvi-EO-2.0-600M & 95.48 & 0.51 & 95.90 & 94.50 \\
 & Prithvi-EO-2.0-600M-TL & 95.43 & 0.25 & 95.70 & 95.00 \\
 & Satlas-Swin-100M & 96.28 & 0.58 & 96.80 & 94.90 \\
 & ScaleMAE-ViT-300M & 91.55 & 1.28 & 93.40 & 89.50 \\
\bottomrule
\end{tabular}
\end{table}

\begin{table}[ht]
\caption{\textcolor{black}{Detailed results for GEO-Bench classification datasets: \textit{m-forestnet}, \textit{m-pv4ger}, and \textit{m-so2sat}.}}
\label{tab:geobench_cls_summary_2}
\begin{tabular}{@{}l@{\hspace{8pt}}rcccc@{}}
\toprule
 & & \multicolumn{4}{c}{Accuracy [\%]} \\
\midrule
dataset & model & mean & std & max & min \\
\midrule
\multirow[t]{13}{45pt}{m-forestnet} & DINO-Resnet50 & 52.47 & 1.40 & 54.18 & 50.45 \\
 & DOFA-ViT-300M & 55.31 & 2.03 & 59.01 & 52.67 \\
 & DeCUR-Resnet50 & 55.90 & 1.22 & 58.61 & 54.18 \\
 & MOCO-Resnet50 & 54.12 & 1.13 & 55.19 & 51.56 \\
 & Prithvi-EO-1.0-100M & 47.89 & 1.05 & 49.55 & 46.02 \\
 & Prithvi-EO-2.0-100M & 51.34 & 2.10 & 53.27 & 47.53 \\
 & Prithvi-EO-2.0-100M-TL & 50.67 & 1.19 & 52.27 & 48.14 \\
 & Prithvi-EO-2.0-300M & 51.74 & 2.06 & 55.49 & 49.35 \\
 & Prithvi-EO-2.0-300M-TL & 53.20 & 1.05 & 55.49 & 51.76 \\
 & Prithvi-EO-2.0-600M & 54.04 & 1.49 & 56.80 & 52.37 \\
 & Prithvi-EO-2.0-600M-TL & 54.53 & 1.84 & 57.91 & 52.27 \\
 & Satlas-Swin-100M & 51.13 & 1.11 & 52.77 & 49.14 \\
 & ScaleMAE-ViT-300M & 45.30 & 1.54 & 47.94 & 43.00 \\
\midrule
\multirow[t]{13}{45pt}{m-pv4ger} & DINO-Resnet50 & 97.47 & 0.34 & 98.00 & 96.80 \\
 & DOFA-ViT-300M & 98.12 & 0.25 & 98.60 & 97.70 \\
 & DeCUR-Resnet50 & 97.38 & 0.30 & 97.90 & 96.90 \\
 & MOCO-Resnet50 & 97.48 & 0.22 & 97.90 & 97.10 \\
 & Prithvi-EO-1.0-100M & 96.50 & 0.25 & 96.90 & 96.00 \\
 & Prithvi-EO-2.0-100M & 97.55 & 0.20 & 97.90 & 97.20 \\
 & Prithvi-EO-2.0-100M-TL & 97.42 & 0.40 & 97.90 & 96.60 \\
 & Prithvi-EO-2.0-300M & 97.88 & 0.37 & 98.40 & 97.30 \\
 & Prithvi-EO-2.0-300M-TL & 97.91 & 0.17 & 98.20 & 97.70 \\
 & Prithvi-EO-2.0-600M & 97.90 & 0.27 & 98.20 & 97.40 \\
 & Prithvi-EO-2.0-600M-TL & 97.80 & 0.24 & 98.20 & 97.40 \\
 & Satlas-Swin-100M & 98.19 & 0.24 & 98.60 & 97.80 \\
 & ScaleMAE-ViT-300M & 96.74 & 0.30 & 97.20 & 96.30 \\
\midrule
\multirow[t]{13}{45pt}{m-so2sat} & DINO-Resnet50 & 54.39 & 1.94 & 56.69 & 51.22 \\
 & DOFA-ViT-300M & 61.26 & 1.97 & 63.89 & 57.81 \\
 & DeCUR-Resnet50 & 56.68 & 1.78 & 58.32 & 53.65 \\
 & MOCO-Resnet50 & 55.58 & 1.87 & 59.74 & 53.25 \\
 & Prithvi-EO-1.0-100M & 51.50 & 1.55 & 54.16 & 49.59 \\
 & Prithvi-EO-2.0-100M & 55.59 & 1.31 & 57.40 & 53.85 \\
 & Prithvi-EO-2.0-100M-TL & 56.80 & 1.28 & 58.22 & 54.67 \\
 & Prithvi-EO-2.0-300M & 58.44 & 1.33 & 60.45 & 56.49 \\
 & Prithvi-EO-2.0-300M-TL & 57.84 & 1.31 & 59.13 & 54.46 \\
 & Prithvi-EO-2.0-600M & 57.20 & 2.17 & 61.46 & 54.16 \\
 & Prithvi-EO-2.0-600M-TL & 59.54 & 2.22 & 64.00 & 55.58 \\
 & Satlas-Swin-100M & 56.74 & 2.70 & 58.82 & 50.51 \\
 & ScaleMAE-ViT-300M & 48.33 & 0.79 & 49.29 & 47.16 \\
\bottomrule
\end{tabular}
\end{table}

\begin{table}[ht]
\caption{\textcolor{black}{Detailed results for GEO-Bench segmentation datasets:  \textit{m-NeonTree}, \textit{m-SA-crop-type}, and \textit{m-cashew-plant}.}}
\label{tab:geobench_seg_summary_1}
\begin{tabular}{@{}l@{\hspace{8pt}}rcccc@{}}
\toprule
 & & \multicolumn{4}{c}{mIoU [\%]} \\
\midrule
dataset & model & mean & std & max & min \\
\midrule
\multirow[t]{13}{45pt}{m-NeonTree} & DINO-Resnet50 & 51.81 & 6.42 & 55.71 & 35.60 \\
 & DOFA-ViT-300M & 58.67 & 1.21 & 61.03 & 56.84 \\
 & DeCUR-Resnet50 & 57.47 & 0.55 & 58.47 & 56.41 \\
 & MOCO-Resnet50 & 56.08 & 0.31 & 56.51 & 55.63 \\
 & Prithvi-EO-1.0-100M & 55.66 & 0.51 & 56.32 & 54.55 \\
 & Prithvi-EO-2.0-100M & 56.57 & 0.49 & 57.44 & 56.02 \\
 & Prithvi-EO-2.0-100M-TL & 56.61 & 0.45 & 57.24 & 55.81 \\
 & Prithvi-EO-2.0-300M & 57.63 & 0.55 & 58.14 & 56.43 \\
 & Prithvi-EO-2.0-300M-TL & 56.95 & 0.61 & 58.13 & 55.89 \\
 & Prithvi-EO-2.0-600M & 59.22 & 0.78 & 60.47 & 58.33 \\
 & Prithvi-EO-2.0-600M-TL & 58.07 & 0.49 & 59.00 & 57.53 \\
 & Satlas-Swin-100M & 57.00 & 0.42 & 57.64 & 56.36 \\
 & ScaleMAE-ViT-300M & 54.85 & 1.36 & 55.89 & 51.38 \\
\midrule
\multirow[t]{13}{45pt}{m-SA-crop-type} & DINO-Resnet50 & 37.26 & 0.75 & 38.34 & 35.99 \\
 & DOFA-ViT-300M & 35.90 & 0.60 & 37.16 & 35.24 \\
 & DeCUR-Resnet50 & 34.49 & 0.63 & 35.36 & 33.15 \\
 & MOCO-Resnet50 & 32.78 & 0.84 & 33.97 & 31.59 \\
 & Prithvi-EO-1.0-100M & 32.50 & 0.73 & 33.37 & 31.07 \\
 & Prithvi-EO-2.0-100M & 38.10 & 0.31 & 38.56 & 37.51 \\
 & Prithvi-EO-2.0-100M-TL & 37.94 & 0.37 & 38.57 & 37.31 \\
 & Prithvi-EO-2.0-300M & 40.38 & 0.24 & 40.78 & 40.05 \\
 & Prithvi-EO-2.0-300M-TL & 40.50 & 0.49 & 41.34 & 39.59 \\
 & Prithvi-EO-2.0-600M & 41.70 & 0.26 & 42.09 & 41.11 \\
 & Prithvi-EO-2.0-600M-TL & 41.36 & 0.35 & 41.85 & 40.60 \\
 & Satlas-Swin-100M & 37.91 & 0.38 & 38.63 & 37.32 \\
 & ScaleMAE-ViT-300M & 25.75 & 0.47 & 26.50 & 25.16 \\
\midrule
\multirow[t]{13}{45pt}{m-cashew-plant} & DINO-Resnet50 & 83.09 & 6.14 & 88.14 & 73.39 \\
 & DOFA-ViT-300M & 81.07 & 6.15 & 88.15 & 74.26 \\
 & DeCUR-Resnet50 & 84.15 & 4.32 & 86.15 & 71.95 \\
 & MOCO-Resnet50 & 78.87 & 6.27 & 85.05 & 70.78 \\
 & Prithvi-EO-1.0-100M & 71.52 & 0.77 & 72.74 & 70.04 \\
 & Prithvi-EO-2.0-100M & 73.42 & 1.05 & 74.72 & 70.91 \\
 & Prithvi-EO-2.0-100M-TL & 78.12 & 5.88 & 83.21 & 71.00 \\
 & Prithvi-EO-2.0-300M & 75.79 & 0.61 & 76.85 & 74.60 \\
 & Prithvi-EO-2.0-300M-TL & 78.47 & 6.12 & 87.66 & 74.12 \\
 & Prithvi-EO-2.0-600M & 85.47 & 5.91 & 90.23 & 76.53 \\
 & Prithvi-EO-2.0-600M-TL & 86.30 & 5.27 & 89.62 & 76.23 \\
 & Satlas-Swin-100M & 73.08 & 6.05 & 82.20 & 68.90 \\
 & ScaleMAE-ViT-300M & 73.60 & 0.98 & 74.56 & 71.34 \\
\bottomrule
\end{tabular}
\end{table}

\begin{table}[ht]
\caption{\textcolor{black}{Detailed results for GEO-Bench segmentation datasets: \textit{m-chesapeake}, \textit{m-nz-cattle}, and \textit{m-pv4ger-seg}.}}
\label{tab:geobench_seg_summary_2}
\begin{tabular}{@{}l@{\hspace{8pt}}rcccc@{}}
\toprule
 & & \multicolumn{4}{c}{mIoU [\%]} \\
\midrule
dataset & model & mean & std & max & min \\
\midrule
\multirow[t]{13}{45pt}{m-chesapeake} & DINO-Resnet50 & 70.81 & 0.64 & 71.46 & 69.80 \\
 & DOFA-ViT-300M & 62.05 & 3.21 & 70.24 & 59.23 \\
 & DeCUR-Resnet50 & 69.83 & 0.61 & 70.80 & 68.86 \\
 & MOCO-Resnet50 & 69.51 & 1.17 & 71.35 & 66.77 \\
 & Prithvi-EO-1.0-100M & 60.18 & 1.99 & 64.01 & 56.83 \\
 & Prithvi-EO-2.0-100M & 59.08 & 1.81 & 62.32 & 57.66 \\
 & Prithvi-EO-2.0-100M-TL & 63.93 & 2.04 & 66.33 & 61.06 \\
 & Prithvi-EO-2.0-300M & 63.42 & 2.35 & 66.69 & 58.91 \\
 & Prithvi-EO-2.0-300M-TL & 65.10 & 1.32 & 67.07 & 63.48 \\
 & Prithvi-EO-2.0-600M & 64.09 & 1.07 & 65.73 & 62.31 \\
 & Prithvi-EO-2.0-600M-TL & 69.19 & 0.84 & 70.73 & 67.72 \\
 & Satlas-Swin-100M & 65.15 & 1.85 & 68.02 & 63.15 \\
 & ScaleMAE-ViT-300M & 53.61 & 1.65 & 56.44 & 50.64 \\
\midrule
\multirow[t]{13}{45pt}{m-nz-cattle} & DINO-Resnet50 & 82.89 & 0.30 & 83.27 & 82.29 \\
 & DOFA-ViT-300M & 77.44 & 0.40 & 78.22 & 76.81 \\
 & DeCUR-Resnet50 & 83.04 & 0.17 & 83.38 & 82.78 \\
 & MOCO-Resnet50 & 81.54 & 0.53 & 82.11 & 80.26 \\
 & Prithvi-EO-1.0-100M & 73.64 & 0.43 & 74.61 & 72.97 \\
 & Prithvi-EO-2.0-100M & 77.21 & 0.30 & 77.64 & 76.62 \\
 & Prithvi-EO-2.0-100M-TL & 77.32 & 0.22 & 77.73 & 76.97 \\
 & Prithvi-EO-2.0-300M & 77.87 & 0.17 & 78.11 & 77.58 \\
 & Prithvi-EO-2.0-300M-TL & 77.47 & 0.34 & 77.91 & 76.78 \\
 & Prithvi-EO-2.0-600M & 81.05 & 0.12 & 81.26 & 80.77 \\
 & Prithvi-EO-2.0-600M-TL & 80.79 & 0.13 & 81.01 & 80.59 \\
 & Satlas-Swin-100M & 81.79 & 0.15 & 81.97 & 81.59 \\
 & ScaleMAE-ViT-300M & 73.71 & 8.76 & 77.01 & 48.82 \\
\midrule
\multirow[t]{13}{45pt}{m-pv4ger-seg} & DINO-Resnet50 & 93.89 & 0.24 & 94.11 & 93.34 \\
 & DOFA-ViT-300M & 94.97 & 0.18 & 95.22 & 94.62 \\
 & DeCUR-Resnet50 & 93.87 & 0.15 & 94.10 & 93.67 \\
 & MOCO-Resnet50 & 94.04 & 0.30 & 94.57 & 93.59 \\
 & Prithvi-EO-1.0-100M & 92.12 & 0.27 & 92.72 & 91.71 \\
 & Prithvi-EO-2.0-100M & 94.47 & 0.18 & 94.76 & 94.15 \\
 & Prithvi-EO-2.0-100M-TL & 94.39 & 0.14 & 94.62 & 94.20 \\
 & Prithvi-EO-2.0-300M & 94.93 & 0.12 & 95.12 & 94.74 \\
 & Prithvi-EO-2.0-300M-TL & 94.88 & 0.15 & 95.08 & 94.55 \\
 & Prithvi-EO-2.0-600M & 95.23 & 0.15 & 95.51 & 95.02 \\
 & Prithvi-EO-2.0-600M-TL & 95.20 & 0.18 & 95.40 & 94.87 \\
 & Satlas-Swin-100M & 95.02 & 0.09 & 95.20 & 94.89 \\
 & ScaleMAE-ViT-300M & 93.56 & 0.23 & 93.85 & 93.16 \\
\bottomrule
\end{tabular}
\end{table}

\end{appendices}

\end{document}